\documentclass[journal]{IEEEtran}
\usepackage{bm}
\usepackage{amsfonts}
\usepackage{amsmath}
\usepackage{booktabs}
\usepackage{multirow}
\usepackage{cite}
\usepackage{array}
\usepackage{pifont}
\usepackage{tikz}
\usepackage{ulem}
\usepackage{hyperref}
\usepackage{subfigure}
\usepackage[table]{xcolor}
\usepackage{xcolor}
\newcommand{\cmark}{{\textcolor{black}{\ding{51}}}}
\newcommand{\xmark}{{\textcolor{lightgray}{\ding{55}}}}
\newcommand{\gtext}[1]{\textcolor[rgb]{0,0.5,0}{#1}}
\newcommand{\rtext}[1]{\textcolor[rgb]{0.8,0,0}{#1}}
\begin{document}
\bstctlcite{IEEEexample:BSTcontrol}
% \title{Bootstrapped Reliable Alignment for Unsupervised Video-Based Visible–Infrared Person Re-Identification}
% Causal Bootstrapped Alignment for Unsupervised Video-Based Visible–Infrared Person Re-Identification
\title{Causal Bootstrapped Alignment for Unsupervised Video-Based Visible–Infrared Person Re-Identification}
\author{Shuang~Li, Jiaxu~Leng, Changjiang~Kuang, Mingpi~Tan, Yu~Yuan and Xinbo Gao, \IEEEmembership{Fellow, IEEE} % <-this % stops a space

% \thanks{This work was supported in part by the National Key R\&D Program of China  under Grant No.2022YFA1004100, in part by the National Natural Science Foundation of China under Grants No. 62472060, U23A20318, and 62221005, in part by the Natural Science Foundation of Chongqing under Grant    No. CSTB2024NSCQ-QCXMX0060, CSTB2023NSCQ-LZX0061, and CSTB2022NSCQ-MSX0547, in part by the Science and Technology Research Program of Chongqing Municipal Education Commission under Grant No. KJZD-K202300604, KJQN202400648, in part by the China Postdoctoral Science Foundation under Grants No. GZC20233362, 2024MD754043, in part by the Chongqing Institute for Brain and Intelligence, in part by Chongqing University of Postsand Telecommunications Ph.D. Innovative Talents Project underGrants BYJS202401. (Corresponding author: J. Leng, X. Gao.)}
\thanks{S. Li, J. Leng, C. Kuang, M. Tan, Y. Yuan and X. Gao,  are with the School of Computer Science and Technology, Chongqing University of Posts and Telecommunications, Chongqing, China (E-mail: shuangli936@gmail.com, lengjx@cqupt.edu.cn, s230201049@stu.cqupt.edu.cn, tanmingp11@163.com, yuyuan9580@gmail.com, gaoxb@cqupt.edu.cn).  (Corresponding author: X. Gao.)}
}

%\markboth{Please submit the manuscript to the Special Issue on Deep Learning for Intelligent Media Computing and Applications}%
%{Shell \MakeLowercase{\textit{et al.}}}
\markboth{Submit to IEEE Transactions on Information Forensics and Security}%
{Shell \MakeLowercase{\textit{et al.}}: Bare Advanced Demo of IEEEtran.cls for IEEE Computer Society Journals}
\maketitle

\begin{abstract}
Video-based visible–infrared person re-identification (VVI-ReID) is a critical technique for all-day surveillance, where temporal information provides additional cues beyond static images. However, existing approaches rely heavily on fully supervised learning with expensive cross-modality annotations, limiting scalability. To address this issue, we investigate Unsupervised Learning for VVI-ReID (USL-VVI-ReID), which learns identity-discriminative representations directly from unlabeled video tracklets. Directly extending image-based USL-VI-ReID methods to this setting with generic pretrained encoders leads to suboptimal performance. Such encoders suffer from weak identity discrimination and strong modality bias, resulting in severe intra-modality identity confusion and pronounced clustering granularity imbalance between visible and infrared modalities. These issues jointly degrade pseudo-label reliability and hinder effective cross-modality alignment. To address these challenges, we propose a Causal Bootstrapped Alignment (CBA) framework that explicitly exploits inherent video priors. First, we introduce Causal Intervention Warm-up (CIW), which performs sequence-level causal interventions by leveraging temporal identity consistency and cross-modality identity consistency to suppress modality- and motion-induced spurious correlations while preserving identity-relevant semantics, yielding cleaner representations for unsupervised clustering. Second, we propose Prototype-Guided Uncertainty Refinement (PGUR), which employs a coarse-to-fine alignment strategy to resolve cross-modality granularity mismatch, reorganizing under-clustered infrared representations under the guidance of reliable visible prototypes with uncertainty-aware supervision. Extensive experiments on the HITSZ-VCM and BUPTCampus benchmarks demonstrate that CBA significantly outperforms existing USL-VI-ReID methods when extended to the USL-VVI-ReID setting.

\end{abstract}
\IEEEpeerreviewmaketitle
\section{Introduction}
\IEEEPARstart{V}{ideo-based} visible-infrared person re-identification (VVI-ReID) \cite{lin2022learning,li2025video,li2023intermediary,du2023video,yang2025dinov2} has emerged as a critical component of all-day surveillance for retrieving pedestrian identities across visible and infrared modalities. 
By leveraging temporal motion cues unavailable in static images \cite{leng2025dual}, VVI-ReID offers superior robustness against occlusion and motion blur.
Unlike conventional person re-identification paradigms that primarily focus on single-modality image-level matching \cite{li2023logical,zhou2026hierarchical,wang2022body}, VVI-ReID introduces cross-modality discrepancies and temporal dynamics, posing additional challenges for representation learning.
However, current approaches rely on fully supervised paradigms that demand exhaustive cross-modality annotations. This dependency is a significant bottleneck, as manually associating identities across disparate modalities is prohibitively expensive and error-prone due to the severe modality gap. To break this dependency, we introduce Unsupervised Learning for VVI-ReID (USL-VVI-ReID). Unlike conventional methods, USL-VVI-ReID learns identity-discriminative representations directly from unlabeled video tracklets, paving the way for scalable and automated surveillance deployments.

\begin{figure}[t!]
\centering
\includegraphics[width=8.7cm,keepaspectratio=true]{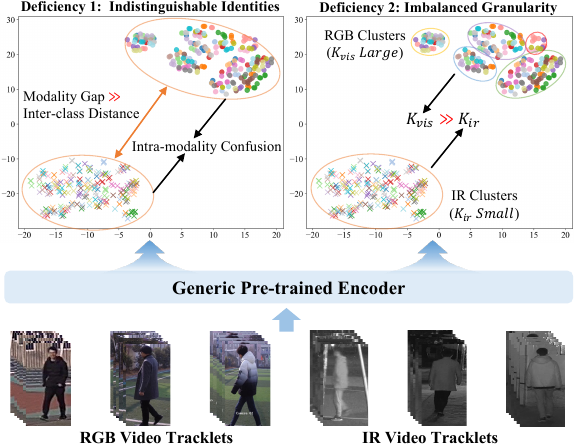}
\caption{Challenges of USL-VVI-ReID with generic pre-trained encoders.
Generic encoders produce two critical deficiencies:
(1) \textbf{indistinguishable identities}, with severe intra-modality confusion and a dominant modality gap; and
(2) \textbf{imbalanced granularity}, where visible features form fine-grained clusters while infrared features are under-clustered, typically resulting in $K_{vis} \gg K_{ir}$ and ambiguous cross-modality associations.}
\label{fig:motivation}
\label{intro1}
\end{figure}

Although image-based unsupervised visible–infrared person re-identification (USL-VI-ReID) \cite{yang2022augmented,wu2023unsupervised,yang2023towards,cheng2023unsupervised,teng2025relieving,yin2024robust} has achieved remarkable progress, directly transferring existing solutions to USL-VVI-ReID remains problematic. Most transfer pipelines rely on a generic pre-trained visual encoder, e.g., CLIP \cite{radford2021learning,yan2023clip,yu2025clip,dong2025diverse}, to extract features, followed by intra-modality clustering to generate pseudo labels and subsequent inter-modality matching for cross-modality alignment. 
However, when directly applied to the video setting, such generic encoders often fail to provide sufficiently identity-discriminative and modality-invariant representations, which leads to two critical deficiencies in the embedding space, as illustrated in Fig.~\ref{fig:motivation}.
First, identity representations become poorly separated, where different individuals are severely entangled within each modality, leading to pronounced intra-modality confusion. Meanwhile, cross-modality discrepancy dominates the feature distribution, causing the modality gap to be substantially larger than inter-class distances.
Second, intra-modality clustering exhibits significant granularity imbalance across modalities.
Specifically, visible features tend to form fine-grained clusters with a large number of discovered clusters $K_{\mathrm{vis}}$,
whereas infrared features remain severely under-clustered with a much smaller $K_{\mathrm{ir}}$.
This imbalance leads to ambiguous cross-modality associations during inter-modality matching,
often manifesting as unstable one-to-many correspondences.

\begin{figure}[t!]
\centering
\includegraphics[width=8.3cm,keepaspectratio=true]{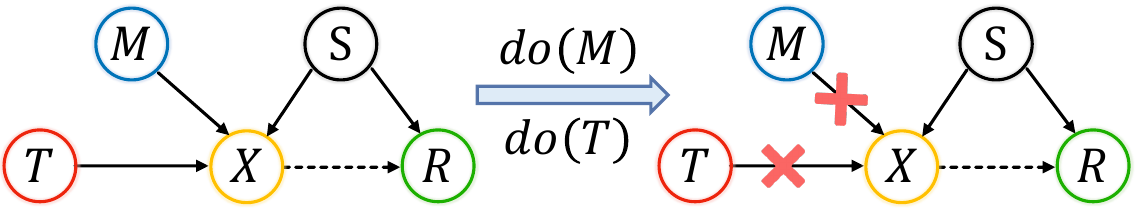}
% \caption{SCM for representation purification.
% The observed input $X$ entangles identity-related factor $S$ and non-causal factors, including modality-specific appearance $M$ and scene-driven temporal cues $T$.
% During representation learning with generic pre-trained models, these non-causal factors may be encoded into the retrieval representation $R$, inducing bias.
% Causal interventions $do(M)$ and $do(T)$ suppress these non-causal influences, yielding an identity-driven and bias-reduced retrieval representation $R$.
% Solid arrows denote directed causal relations, while dashed arrows indicate statistical associations.}
\caption{Structural causal model (SCM) for representation purification.
The observed input $X$ entangles identity factor $S$ and non-causal factors, including modality-specific appearance $M$ and scene-driven temporal cues $T$.
During representation learning, these non-causal factors may be encoded into the retrieval representation $R$.
Causal interventions $do(M)$ and $do(T)$ aim to suppress these influences.
Solid arrows denote directed causal relations, while dashed arrows indicate statistical associations.}
\label{fig:causal_graph}
\end{figure}

The above identity confusion issue stems from the intrinsic behavior of generic pre-trained encoders under unsupervised training.
Being task-agnostic by design, such encoders indiscriminately encode all predictable signals in video sequences, resulting in representations that entangle identity-related and non-causal factors.
As illustrated in Fig.~\ref{fig:causal_graph}, we formalize this process using a structural causal model (SCM)~\cite{tang2025boosting,liu2024causality,lv2022causality}.
The observed input $X$ jointly contains the intrinsic identity factor $S$ and two non-causal confounders:
modality-specific appearance $M$, capturing modality-dependent spectral statistics (e.g., color in visible videos and thermal textures in infrared videos),
and scene-driven temporal cues $T$, encoding camera- and scene-dependent motion regularities (e.g., walking direction, camera viewpoint bias, and background motion patterns).
Without identity supervision, $M$ and $T$ form backdoor paths to the learned representation $R$, inducing spurious correlations that dominate the embedding space.
Such representation bias arises before clustering, and directly applying clustering on these confounded features inevitably amplifies spurious correlations through noisy pseudo labels, underscoring the necessity of calibrating generic pre-trained encoders before clustering-based learning.
However, eliminating non-causal factors alone is insufficient for reliable identity discovery.
Overly aggressive causal interventions may suppress identity-discriminative cues, yielding representations that are invariant yet insufficient.
From a causal perspective, an ideal representation should simultaneously satisfy
\emph{Separation}, by blocking spurious backdoor paths, and
\emph{Causal Sufficiency}, by preserving task-relevant identity semantics~\cite{lv2022causality}. In the video setting, causal sufficiency can be naturally supported by inherent identity consistency within video sequences, providing a crucial prior for preserving identity-discriminative information under unsupervised learning.

Moreover, causal calibration alone cannot eliminate the inherent asymmetry in discriminative capacity across modalities. This asymmetry arises from the unequal availability of identity cues in video sequences: visible sequences preserve richer appearance details and more stable spatio-temporal structures that support fine-grained discrimination, whereas infrared sequences provide fewer discriminative cues and thus exhibit intrinsically coarser structures \cite{ye2021deep,ye2020dynamic,wu2021discover}.
Consequently, even with causally purified representations, the infrared modality tends to form fewer and coarser clusters than the visible modality, making it difficult to establish reliable one-to-one correspondences between visible and infrared clusters \cite{wu2023unsupervised}. This observation motivates leveraging cross-modality complementarity, where the more discriminative visible modality serves as a structural reference to guide refinement of the infrared representation space.

Based on the above analysis, we propose Causal Bootstrapped Alignment (CBA), an unsupervised video-based visible–infrared person re-identification (USL-VVI-ReID) framework designed to overcome the adaptation limitations of generic pre-trained models through two complementary components: Causal Intervention Warm-up (CIW) and Prototype-Guided Uncertainty Refinement (PGUR).
Specifically, CIW performs sequence-level causal interventions on modality and temporal factors by constructing counterfactual representations to calibrate the encoder. Modality-specific appearance is intervened via cross-modality style transfer to disrupt spectral statistics, while scene-driven temporal cues are intervened through random frame swapping within sequences to break spurious motion regularities. 
By enforcing representation consistency between original and intervened sequences, CIW suppresses non-causal shortcuts while preserving sequence-level identity semantics, yielding identity-focused representations with a cleaner feature space for subsequent learning.
To further address modality-induced clustering granularity imbalance, PGUR conducts structure refinement during cross-modal matching. It leverages discriminative visible prototypes as reference anchors to guide refinement of the intrinsically coarser infrared representation space. PGUR first establishes prototype-level cross-modal associations via iterative graph matching. Based on the resulting matching structures, one-to-one correspondences are treated as reliable associations and optimized with hard contrastive supervision, while one-to-many correspondences are regarded as ambiguous and handled with soft contrastive supervision to prevent error propagation. Through this uncertainty-aware refinement, PGUR progressively transfers clustering granularity from the visible to the infrared modality, enabling robust cross-modal correspondence and modality-invariant representation learning.

Our main contributions are summarized as follows:
\begin{itemize}
    \item We propose CBA, a progressive framework for USL-VVI-ReID, which integrates causal representation calibration and cross-modality association refinement, and to the best of our knowledge, constitutes the first exploration of USL-VVI-ReID.
    \item We introduce a Causal Intervention Warm-up (CIW) that performs sequence-level causal interventions to calibrate generic pre-trained encoders toward identity-focused representations, enabling reliable unsupervised clustering.
    \item We introduce a Prototype-Guided Uncertainty Refinement (PGUR) to address modality-induced clustering granularity mismatch by exploiting cross-modality complementarity, where discriminative visible prototypes serve as reference anchors for uncertainty-aware refinement of infrared representations.
    \item Extensive experiments on HITSZ-VCM and BUPTCampus demonstrate that CBA significantly outperforms existing USL-VI-ReID methods when extended to the USL-VVI-ReID setting, establishing state-of-the-art performance on this task.
\end{itemize}

The rest of this paper is organized as follows. Section \uppercase\expandafter{\romannumeral2} introduces related work; Section \uppercase\expandafter{\romannumeral3}  elaborates the proposed method; Section \uppercase\expandafter{\romannumeral4} analyzes the comparative experimental results; and Section \uppercase\expandafter{\romannumeral5} concludes this paper.

\section{Related Work}
\subsection{Video-based Visible-Infrared Person Re-Identification}
Video-based visible–infrared person re-identification (VVI-ReID) extends image-based retrieval \cite{li2025shape} by leveraging temporal cues to enhance discrimination and mitigate cross-modality discrepancies. Early studies primarily focused on temporal aggregation to model motion evolution, such as MITML~\cite{lin2022learning}, which introduced modal-invariant temporal memory learning. To overcome the limited long-range modeling capability of CNNs, CST~\cite{feng2024cross} adopted a cross-modality spatial–temporal transformer, while IBAN~\cite{li2023intermediary} utilized an intermediary modality for bidirectional aggregation.
Recent efforts have further emphasized data scale, robustness, and structured representation learning. AuxNet~\cite{du2023video} constructed the large-scale BUPTCampus dataset with curriculum-based auxiliary learning, and SAADG~\cite{zhou2023video} alleviated intra-modal variations via style augmentation and graph-based interaction. Meanwhile, multi-granularity modeling and unified representations have gained attention: HD-GI~\cite{zhou2025hierarchical} captured invariant partial cues through hierarchical disturbance, while CLIP-based approaches, such as VLD~\cite{li2025video} and X-ReID~\cite{yu2025x}, enhanced cross-modality alignment via language-driven prompting and prototype collaboration.
Despite these advances, most existing VVI-ReID methods still depend on costly cross-modality identity annotations, limiting scalability to large-scale video data. This challenge motivates Unsupervised Video-based Visible–Infrared Person Re-Identification (USL-VVI-ReID), which aims to learn robust video representations without identity supervision.

\subsection{Unsupervised Video Person Re-Identification}
Unsupervised video person re-identification seeks to learn discriminative video representations without identity annotations, with early studies largely relying on iterative pseudo-label estimation for unlabeled tracklets. Representative methods, such as Stepwise~\cite{liu2017stepwise} and RACE~\cite{ye2018robust}, refined cross-camera associations via metric promotion and anchor-based graph construction, while DGM~\cite{ye2019dynamic} and CCM~\cite{wang2020exploiting} further improved label reliability through graph co-matching and global camera constraints. In parallel, association-driven frameworks, including TAUDL~\cite{li2018unsupervised}, UTAL~\cite{li2019unsupervised}, and TS-DTW~\cite{ma2017person}, emphasized joint optimization of within- and cross-camera tracklet alignment. To mitigate noise and frame-level variations, later works introduced sampling, anchor association, and progressive learning strategies~\cite{xie2022sampling,zeng2022anchor,fan2018unsupervised}.
However, these methods are primarily designed for single-modality scenarios and implicitly assume appearance consistency, making them inadequate for VVI-ReID, where the severe visible–infrared modality gap induces substantial representation bias.

\subsection{Unsupervised Image-based Visible-Infrared Person Re-identification}
Unsupervised visible–infrared person re-identification aims to learn modality-invariant representations without identity supervision, typically relying on clustering-based pseudo-label estimation and cross-modality alignment. Early studies mainly focused on alleviating modality discrepancies and mitigating pseudo-label noise. ADCA~\cite{yang2022augmented} introduced a dual-contrastive aggregation framework with style augmentation to reduce intra- and inter-modality gaps, while DOTLA~\cite{cheng2023unsupervised}, N-ULC~\cite{teng2025relieving}, and RDL~\cite{li2025robust} refined pseudo-label reliability through neighbor-guided or centroid-aware calibration strategies.
Beyond clustering, more advanced association mechanisms have been explored to improve cross-modality matching. PGM~\cite{wu2023unsupervised} modeled correspondence through progressive graph matching, and M$^3$\cite{shi2024multi} captured diverse intra-class variations via multi-memory matching. To handle hard samples and outliers, CSANet\cite{xi2025csanet} and DLM~\cite{ye2025dual} adopted self-paced and dual-level association strategies with explicit filtering. In addition, SDCL~\cite{yang2024shallow} enforced consistency between shallow texture and deep semantic features, and UP-CLIP~\cite{chen2023unveiling} leveraged vision–language alignment to guide unsupervised representation learning.
Nevertheless, existing methods remain image-centric and fail to exploit temporal continuity, which limits their ability to maintain identity consistency in USL-VVI-ReID under severe modality discrepancies.

\section{Proposed method}
In this section, we introduce the proposed USL-VVI-ReID framework (Fig.~\ref{framework}),
which consists of Causal Intervention Warm-up (CIW) and Prototype-Guided Uncertainty
Refinement (PGUR). We then present the associated optimization objectives.

\subsection{Preliminary}\label{sec3.1}
USL-VVI-ReID aims to learn modality-invariant sequence-level representations from unlabeled visible–infrared video tracklets via iterative pseudo-label optimization. Building upon USL-VI-ReID frameworks such as ADCA \cite{yang2022augmented}, we extend this paradigm to the video-based cross-modality setting. For stable sequence representation learning, we adopt the supervised VVI-ReID model VLD \cite{li2025video} as the backbone, which explicitly captures inter-frame interactions.

\begin{figure*}[t!]
\centering
\includegraphics[width=17.5cm,keepaspectratio=true]{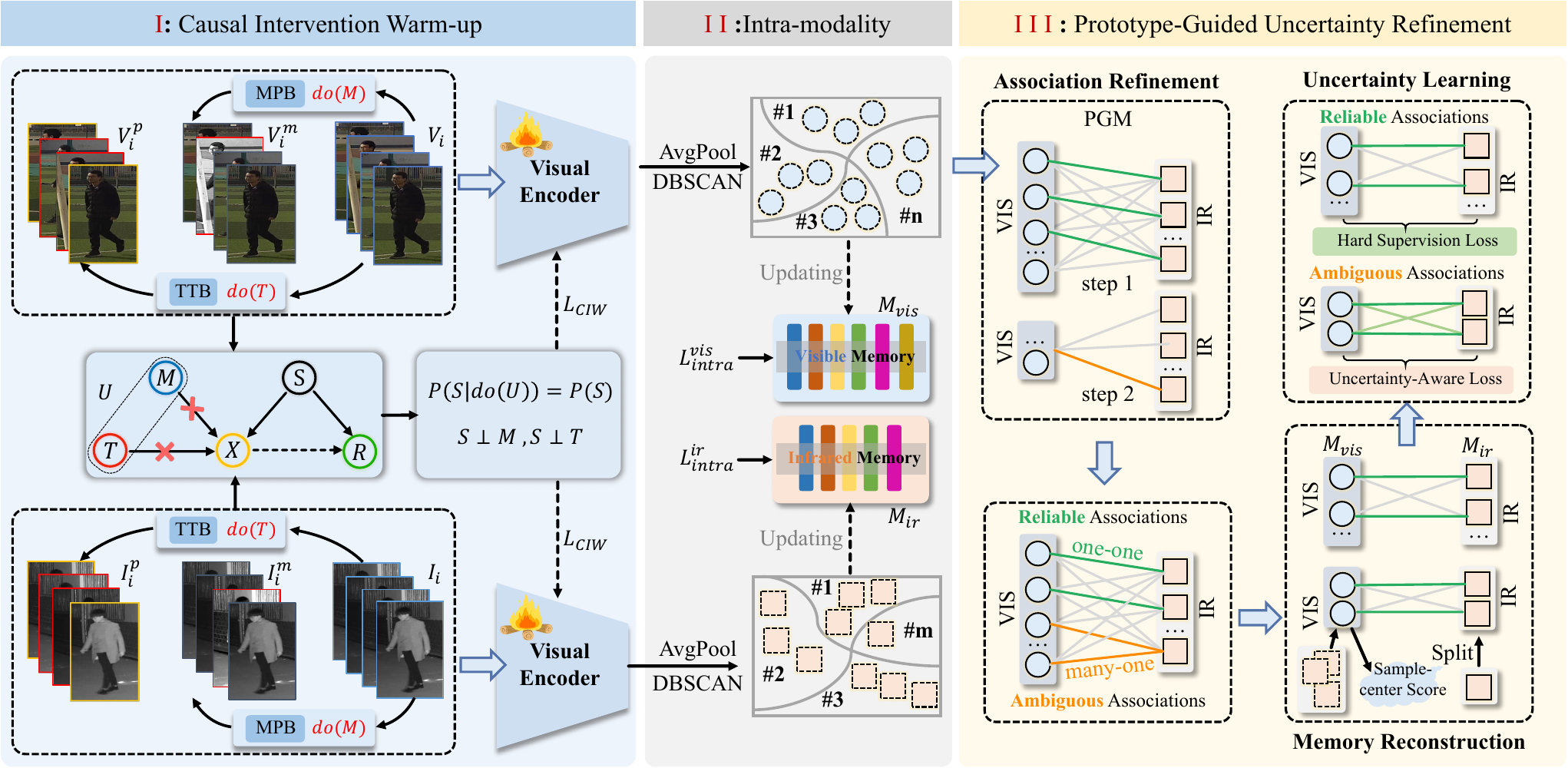}
\caption{Overall architecture of the proposed CBA framework for USL-VVI-ReID.
The framework comprises three stages:
\textbf{(I) Causal Intervention Warm-up (CIW)}, which calibrates the encoder via sequence-level causal intervention to mitigate modality and temporal confounders while preserving identity-related information;
\textbf{(II) Intra-modality Learning}, which performs clustering and intra-modality contrastive learning;
and \textbf{(III) Prototype-Guided Uncertainty Refinement (PGUR)}, which resolves the granularity mismatch through uncertainty-aware cross-modality refinement guided by reliable visible prototypes.}
\label{framework}
\end{figure*}

As shown in Fig.\ref{framework}, visible and infrared video sequences
$V=\{x^{t}_{vis}\}_{t=1}^{T}$ and $I=\{x^{t}_{ir}\}_{t=1}^{T}$ are encoded by a shared visual encoder $E_v(\cdot)$\cite{li2025video}.
For clarity, we use the visible sequence as an example.
The encoder $E_v$ jointly models all frames via STP~\cite{li2025video} to capture spatio-temporal dependencies, yielding temporally enhanced frame-level features:
\begin{equation}
\{\mathbf{f}_{vis}^{1}, \mathbf{f}_{vis}^{2}, \dots, \mathbf{f}_{vis}^{T}\} = E_{v}(V),
\end{equation}
where $\mathbf{f}^{t}_{vis} \in \mathbb{R}^d$ denotes the temporally enhanced feature of the $t$-th frame.
Then, temporal average pooling $\text{AvgPool}$ is applied across these representations to obtain the sequence-level feature:
\begin{equation}
\mathbf{f}_{vis} = \text{AvgPool}([\mathbf{f}_{vis}^{1}, \mathbf{f}_{vis}^{2}, \dots, \mathbf{f}_{vis}^{T}]).
\end{equation}
We denote the set of sequence-level features for all visible tracklets as
$\mathcal{F}_{vis}=\{\mathbf{f}_{vis,i}\}_{i=1}^{N_{vis}}$, where $N_{vis}$ is the number of visible sequences.
Pseudo labels are then generated by applying DBSCAN clustering to $\mathcal{F}_{vis}$:
\begin{equation}
\hat{\mathcal{Y}}_{vis} = \text{DBSCAN}(\mathcal{F}_{vis}),
\end{equation}
where $\hat{\mathcal{Y}}_{vis} = \{\hat{y}_{vis,i}\}_{i=1}^{N_{vis}}$ represents the pseudo labels for all visible sequences, and $\hat{y}_{vis,i} \in \{0,1,\dots,K_{vis}-1\}$ denotes the pseudo class label assigned to the $i$-th visible sequence, with $K_{vis}$ being the number of clusters discovered by DBSCAN.

Subsequently, a visible memory bank $\mathcal{M}_{vis}=\{\mathbf{m}_{vis}^{c}\}_{c=0}^{K_{vis}-1}$ is then constructed to store cluster prototypes, where each prototype $\mathbf{m}_{vis}^{c}$ is initialized as the centroid of visible sequences assigned to cluster $c$:
\begin{equation}
\mathbf{m}_{vis}^{c} = \frac{1}{N_{vis}^{c}} \sum_{i=1}^{N_{vis}} \mathbb{I}(\  \hat{y}_{vis,i} = c) \cdot \mathbf{f}_{vis,i},
\end{equation}
where $N_{vis}^{c}$ denotes the number of visible sequences in class $c$, and $\mathbb{I}(\cdot)$ is the indicator function that returns 1 if the condition is true and 0 otherwise.

To enhance intra-modality discriminability, the anchor sequence feature $\mathbf{f}_{vis,i}$ is pulled toward its corresponding prototype $\mathbf{m}_{vis}^{+}$ within the memory bank. The intra-modality contrastive loss is formulated as follows: 
\begin{equation}
\mathcal{L}_{intra}^{vis} =
- \log
\frac{\exp(\mathbf{f}_{vis,i} \cdot \mathbf{m}_{vis}^{+}/\tau)}
{\sum_{c=1}^{K_{vis}}\exp(\mathbf{f}_{vis,i} \cdot \mathbf{m}_{vis}^{c}/\tau)},
\end{equation}
where $\tau$ is the temperature coefficient. 
An identical intra-modality loss $\mathcal{L}_{intra}^{ir}$ is symmetrically applied to the infrared modality using the infrared memory bank $\mathcal{M}_{ir}$. The total intra-modality optimization objective for this stage is formulated as the joint loss of both modalities:
\begin{equation}
\mathcal{L}_{intra} = \mathcal{L}_{intra}^{vis} + \mathcal{L}_{intra}^{ir},
\end{equation}
which establishes a stable and discriminative intra-modality representation space, facilitating subsequent cross-modality alignment.

% #############################

% #############################
\subsection{Causal Intervention Warm-up}\label{sec3.2}
Existing USL-VI-ReID methods typically perform clustering on features extracted directly from pretrained models. 
However, in the cross-modality video scenario, such features are heavily confounded by modality-specific styles and unstable motion patterns. 
Without identity guidance, the initial encoder tends to capture these non-causal shortcuts, leading to unreliable clusters and severe error accumulation during the subsequent iterative optimization. 
To address this ``cold-start'' challenge, we propose the Causal Intervention Warm-up (CIW),
a crucial pre-clustering calibration stage that leverages causal interventions to
purify representations and emphasize intrinsic identity cues.

\textbf{Structural Causal Model and Causal Properties.}
Inspired by CIRL~\cite{lv2022causality}, we formalize the video representation generation process as a Structural Causal Model (SCM).
As illustrated in Fig.~\ref{fig:causal_graph}, an observed video tracklet $X$ is modeled as being generated from both causal and non-causal factors:
\begin{equation}
    X := f(S, U, \epsilon),
\end{equation}
where $f(\cdot)$ denotes an unknown causal generation mechanism that maps latent
factors to the observed video, $S$ denotes the causal factors representing identity-related features that remain invariant to modality and temporal variations, $U = \{M, T\}$ represents the non-causal confounders including modality-specific style $M$ and temporal topology $T$, and $\epsilon$ signifies the unexplained noise variables \cite{lv2022causality}. The primary objective of causal representation learning is to extract these latent causal factors $S$ from the observed $X$ to reconstruct an invariant causal mechanism for person identification.

\begin{figure}[t!]
\centering
\includegraphics[width=8.5cm,keepaspectratio=true]{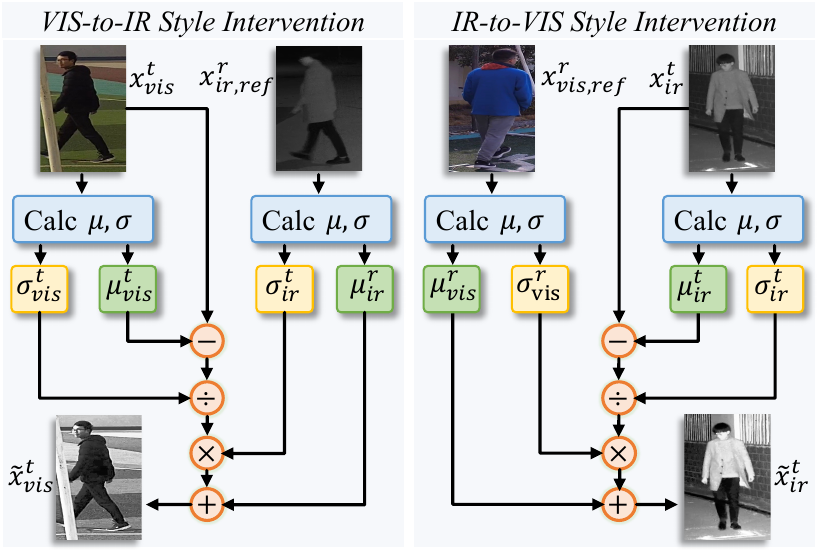}
\caption{Illustration of the Modality-Perturbation Bootstrapping (MPB).
MPB applies bidirectional VIS$\leftrightarrow$IR style interventions by transferring global appearance statistics (mean $\mu$, std $\sigma$) between modalities, producing modality-perturbed samples for representation calibration.}
\label{fig:mpb}
\end{figure}

Without explicit identity supervision, generic encoders tend to capture spurious statistical correlations rather than the underlying causal mechanism.
This induces spurious backdoor paths (e.g., $M \to X \to R$ and $T \to X \to R$),
causing the model to rely on modality-specific spectral noise or unstable motion
patterns for identity discrimination.
To block these paths and recover the causal factors $S$, the learned representations
are required to satisfy the properties of ideal causal mechanisms.
According to the Independent Causal Mechanisms (ICM) principle
\cite{peters2017elements,scholkopf2012causal}, an effective representation should
possess:
\begin{itemize}
    \item \textbf{Causal Separation:} The causal factor $S$ should be independent
    of non-causal factors $U$, i.e., $S \perp U$, such that interventions on $U$ do not
    alter $S$, satisfying $P(S \mid do(U)) = P(S)$.
\item \textbf{Causal Sufficiency:} The representation should retain all causal information necessary for reliable identity discrimination in person re-identification.
\end{itemize}

Since the intrinsic causal factor $S$ is latent and cannot be directly recovered from raw video data, we instead learn causal representations by enforcing the aforementioned properties. Specifically, to satisfy \emph{Causal Separation}, we simulate the $do$-operator by intervening on non-causal factors while preserving identity semantics.
We introduce complementary interventions $do(M)$ and $do(T)$ to eliminate modality and temporal confounders, together with a stabilization mechanism to ensure \emph{Causal Sufficiency}.

\textbf{Modality-Perturbation Bootstrapping.}
To implement the modality-level intervention $do(M)$ and enforce modality independence $S \perp M$, we propose Modality-Perturbation Bootstrapping (MPB).
MPB mitigates modality-specific confounding by perturbing appearance cues that stem from substantial cross-modality discrepancies, such as illumination and intensity statistics.
Specifically, given a sequence $V_i=\{x_i^t\}_{t=1}^T$, MPB randomly perturbs the appearance of a subset of frames rather than the entire tracklet, thereby disrupting modality-dependent signals while preserving identity semantics through temporal context.
As illustrated in Fig.~\ref{fig:mpb}, for each target frame $x_i^t$, a reference frame $x_{ref}^r$ is sampled from the opposite modality as the style source.
Inspired by AdaIN~\cite{huang2017arbitrary}, we inject the global style statistics of $x_{ref}^r$ into $x_i^t$ via cross-modality style transfer:
\begin{equation}
\tilde{x}_{i}^{t} = \sigma(x_{ref}^{r}) \cdot \frac{x_{i}^{t} - \mu(x_{i}^{t})}{\sigma(x_{i}^{t})} + \mu(x_{ref}^{r}),
\end{equation}
where $\mu(\cdot)$ and $\sigma(\cdot)$ denote the channel-wise mean and standard deviation.
Under this formulation, $x_{i}^{t}$ preserves identity-related content, while
$x_{ref}^{r}$ supplies modality-specific style statistics for the interventional
perturbation.

To bridge frame-level perturbations with sequence-level representation learning,
we construct a counterfactual modality view $V_{i}^{m}$ as a hybrid sequence.
Let $\mathcal{T}_{sub} \subset \{1, \dots, T\}$ denote the indices of frames randomly
selected for intervention.
The resulting sequence $V_{i}^{m} = \{\hat{x}_{i}^{t}\}_{t=1}^{T}$ is defined as:
\begin{equation}
    \hat{x}_{i}^{t} = 
    \begin{cases} 
        \tilde{x}_{i}^{t}, & \text{if } t \in \mathcal{T}_{sub} \\
        x_{i}^{t}, & \text{if } t \notin \mathcal{T}_{sub}
    \end{cases}
\end{equation}
where $\tilde{x}_{i}^{t}$ is the style-perturbed frame. By retaining a portion of original frames while injecting modality noise into others, $V_{i}^{m}$ preserves the identity-related temporal anchors while explicitly disrupting the global modality bias.

Subsequently, we feed both the original and counterfactual sequences into the encoder to obtain sequence-level representations: $\mathbf{f}_{i} = \text{AvgPool}(E_{v}(V_{i}))$ and $\mathbf{f}_{i}^{m} = \text{AvgPool}(E_{v}(V_{i}^{m}))$. To satisfy the Separation property, we define the original-to-counterfactual contrastive loss $\mathcal{L}_{o2c}^{m}$ and the counterfactual-to-original contrastive loss $\mathcal{L}_{c2o}^{m}$ as:
\begin{equation}
    \begin{aligned}
\mathcal{L}_{o2c}^{m} &= -\frac{1}{B} \sum_{i=1}^{B} \log \frac{\exp(\mathbf{f}_{i} \cdot \mathbf{f}_{i}^{m} / \tau)}{\sum_{j=1}^{B} \exp(\mathbf{f}_{i} \cdot \mathbf{f}_{j}^{m} / \tau)},\\
\mathcal{L}_{c2o}^{m} &= -\frac{1}{B} \sum_{i=1}^{B} \log \frac{\exp(\mathbf{f}_{i}^{m} \cdot \mathbf{f}_{i} / \tau)}{\sum_{j=1}^{B} \exp(\mathbf{f}_{i}^{m} \cdot \mathbf{f}_{j} / \tau)},
    \end{aligned}
\end{equation}
where $B$ denotes the mini-batch size. The total modality intervention loss $\mathcal{L}_{\mathrm{mpb}}$ is calculated as follows: 
\begin{equation} \mathcal{L}_{mpb} = \frac{1}{2} (\mathcal{L}_{o2c}^{m} + \mathcal{L}_{c2o}^{m}), 
\end{equation} 
which encourages balanced alignment between the original and counterfactual domains, thereby suppressing non-causal modality-specific statistics in the learned representations.

\textbf{Temporal-Topology Bootstrapping.}
To mitigate the influence of spurious temporal shortcuts and intervene on temporal confounders $T$, we propose Temporal-Topology Bootstrapping (TTB) to approximate the intervention $do(T)$.
TTB targets the tendency of generic pre-trained encoders to exploit stable temporal ordering as motion-based shortcuts, rather than modeling identity-consistent dynamics.
Specifically, given a video sequence $V_i$, we construct a counterfactual temporal view $V_i^{p}$ by randomly swapping adjacent frames, thereby perturbing the original temporal topology while preserving the underlying identity content.
The perturbed sequence is then encoded to obtain a sequence-level representation
$\mathbf{f}_i^{p} = \text{AvgPool}(E_{v}(V_i^{p}))$.
Similar to MPB, we employ original-to-counterfactual and counterfactual-to-original temporal contrastive losses,
$\mathcal{L}_{o2c}^{s}$ and $\mathcal{L}_{c2o}^{s}$,
to suppress spurious temporal dependencies and encourage temporally invariant identity representations:
\begin{equation}
    \begin{aligned}
\mathcal{L}_{o2c}^{p} &= -\frac{1}{B} \sum_{i=1}^{B} \log \frac{\exp(\mathbf{f}_{i} \cdot \mathbf{f}_i^{p} / \tau)}{\sum_{j=1}^{B} \exp(\mathbf{f}_{i} \cdot \mathbf{f}_{j}^{p} / \tau)},\\
% \end{equation}
% \begin{equation}
\mathcal{L}_{c2o}^{p} &= -\frac{1}{B} \sum_{i=1}^{B} \log \frac{\exp(\mathbf{f}_i^{p} \cdot \mathbf{f}_{i} / \tau)}{\sum_{j=1}^{B} \exp(\mathbf{f}_i^{p} \cdot \mathbf{f}_{j} / \tau)}.
    \end{aligned}
\end{equation}
The total temporal intervention loss $\mathcal{L}_{\mathrm{ttb}}$ is then formulated as:
\begin{equation}
\mathcal{L}_{ttb} = \frac{1}{2} (\mathcal{L}_{o2c}^{p} + \mathcal{L}_{c2o}^{p}).
\end{equation}
By minimizing $\mathcal{L}_{ttb}$, the model is encouraged to learn representations that are invariant to temporal-topological disruptions, thereby suppressing motion-based shortcuts and emphasizing stable identity-related semantics.

\textbf{Identity-Consistency Stabilization.}
While MPB and TTB effectively promote the \emph{Causal Separation} property by eliminating
non-causal modality and temporal shortcuts, overly aggressive interventional de-biasing may inadvertently attenuate identity-discriminative cues.
To ensure \emph{Causal Sufficiency}, we introduce Identity-Consistency Stabilization (ICS),
which provides a stable identity constraint during representation calibration.
ICS leverages a fundamental video prior that frames within the same tracklet share
a consistent identity despite pose, and enforces the stability of the latent identity factor $S$ across the sequence.
As the causal factor underlying identity discrimination, $S$ should remain stable
throughout a tracklet; accordingly, ICS acts as a functional constraint that preserves
task-relevant identity information in the mapping from $X$ to $R$, preventing the
interventional de-biasing process from discarding discriminative causal signals.

Specifically, given a video sequence $V_i$, the visual encoder extracts a set of frame-level features
$\mathcal{F}_{i}^{seq}=\{\mathbf{f}^{t}_{i}=E_{v}(x^{t}_{\mathrm{vis},i})\}_{t=1}^{T}$.
An anchor feature $\mathbf{f}_{i}^{a}$ is randomly sampled from $\mathcal{F}_{i}^{seq}$.
The remaining frames within the same sequence constitute the positive set $\mathcal{P}_{i}$,
while frames from other sequences form the negative set $\mathcal{N}_{i}$.
Since video sequences are randomly sampled in each mini-batch, the probability that two
sequences share the same identity is negligible, ensuring that negative samples reliably
correspond to identity changes.
Based on this construction, we enforce an identity-consistency constraint using a
sequence-consistency loss inspired by the weighted regularized triplet (WRT)
\cite{ye2021channel}:
\begin{equation}
\mathcal{L}_{seq}
=\frac{1}{B}\sum_{i=1}^{B}
\log\Bigl(1+\exp\bigl(
\sum_{\mathbf{f}^{p}_{i}\in P_{i}} w^{p}_{i,p}\, d^{p}_{i,p}
-
\sum_{\mathbf{f}^{n}_{j}\in N_{i}} w^{n}_{i,j}\, d^{n}_{i,j}
\bigr)\Bigr),
\end{equation}
where $B$ is the batch size, the distance between the anchor and a positive
sample is computed as $d^{p}_{i,p}=\|\mathbf{f}^{a}_{i}-\mathbf{f}^{p}_{i}\|_{2}$,
and the distance between the anchor and a negative sample is
$d^{n}_{i,j}=\|\mathbf{f}^{a}_{i}-\mathbf{f}^{n}_{j}\|_{2}$. The associated
weights are obtained through softmax normalization, where each positive sample
is assigned 
$w^{p}_{i,p}=\exp(d^{p}_{i,p})\big/\sum_{\mathbf{f}^{p}_{i}\in P_{i}}\exp(d^{p}_{i,p})$,
and each negative sample is assigned 
$w^{n}_{i,j}=\exp(-d^{n}_{i,j})\big/\sum_{\mathbf{f}^{n}_{j}\in N_{i}}\exp(-d^{n}_{i,j})$.

\subsection{Prototype-Guided Uncertainty Refinement}
\label{sec:PGUR}
\begin{figure}[t!]
\centering
\includegraphics[width=8.9cm,keepaspectratio=true]{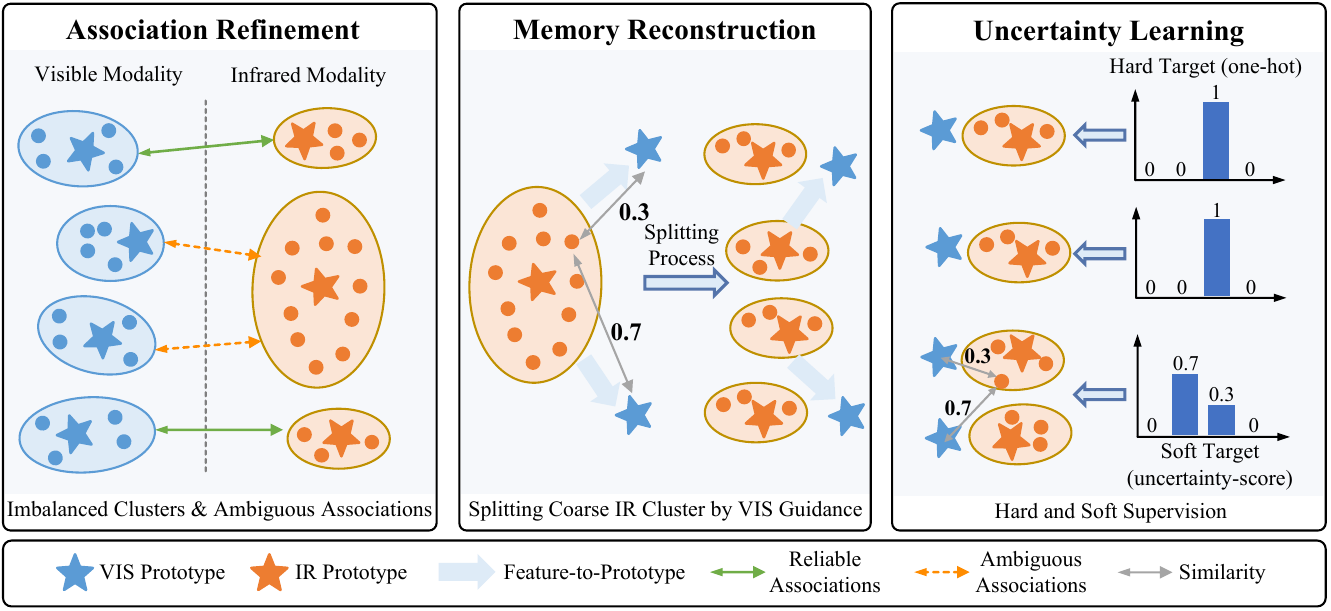}
\caption{Illustration of the PGUR.
PGUR addresses cross-modality granularity mismatch by first separating reliable one-to-one and ambiguous one-to-many associations (left),
then reconstructing a structurally aligned prototype memory by splitting coarse infrared clusters under visible guidance (middle),
and finally applying hard supervision to reliable pairs and uncertainty-aware soft supervision to ambiguous ones for robust cross-modality alignment (right).}
\label{PGUR}
\end{figure}

Although Causal Intervention Warm-up (CIW) establishes a purified representation basis by satisfying causal separation and causal sufficiency, it does not inherently resolve the cross-modality association problem. After performing intra-modality clustering, the visible and infrared feature spaces still exhibit inconsistent structural patterns. In practice, the visible modality typically forms a significantly larger number of clusters than the infrared modality ($K_{\mathrm{vis}} \gg K_{\mathrm{ir}}$). This granularity mismatch reflects the inherent asymmetry in discriminative capacity: visible sequences preserve richer appearance details, whereas infrared sequences provide limited texture cues, leading to an under-clustering effect in the infrared domain. Such imbalance makes cross-modality association inherently difficult, as bipartite matching often yields ambiguous one-to-many correspondences rather than reliable one-to-one identities \cite{wu2023unsupervised}.
To address this structural inconsistency, we introduce Prototype-Guided Uncertainty Refinement (PGUR). PGUR leverages the more discriminative visible prototypes as structural anchors to guide the refinement of the intrinsically coarser infrared representation space. As illustrated in Fig.~\ref{PGUR}, the process consists of three key steps: \textbf{Prototype-Guided Association Refinement}, \textbf{Structural Memory Reconstruction}, and \textbf{Uncertainty-Aware Contrastive Learning}.

\textbf{Association Refinement.}
Following~\cite{wu2023unsupervised}, we adopt a progressive bipartite graph matching
scheme to establish cross-modality associations between visible cluster prototypes
$\mathcal{M}_{vis}=\{\mathbf{m}_{vis}^{i}\}_{i=1}^{K_{vis}}$
and infrared cluster prototypes
$\mathcal{M}_{ir}=\{\mathbf{m}_{ir}^{j}\}_{j=1}^{K_{ir}}$.
Each matching step is formulated as:
\begin{equation}
\begin{aligned}
\min_{\mathbf{M}} \quad &
\sum_{i=1}^{K_{vis}}\sum_{j=1}^{K_{ir}} 
C(i,j)\,\mathbf{M}_{ij} \\
\text{s.t.}\quad &
\mathbf{M}_{ij}\in\{0,1\}, \quad
\sum_{j}\mathbf{M}_{ij}\le 1, \quad
\sum_{i}\mathbf{M}_{ij}=1,
\end{aligned}
\end{equation}
where $C(i,j)=1-\operatorname{Sim}(\mathbf{m}_{vis}^{i},\mathbf{m}_{ir}^{j})$ denotes
the cross-modality dissimilarity based on cosine similarity, and
$\mathbf{M}\in\{0,1\}^{K_{vis}\times K_{ir}}$ is the assignment matrix.

Due to the intrinsic granularity imbalance ($K_{vis} > K_{ir}$), progressive
matching naturally yields one-to-many correspondence patterns, where a single
infrared prototype may be associated with multiple visible prototypes.
Based on the matching results, for each infrared cluster $\mathbf{m}_{ir}^{j}$,
we define its associated visible index set as:
\begin{equation}
\mathcal{V}_{j}
=\{\, i \mid (i,j)\in\mathcal{M}^{(1)}\cup\mathcal{M}^{(2)} \,\},
\end{equation}
where $\mathcal{M}^{(1)}$ and $\mathcal{M}^{(2)}$ denote the sets of visible–infrared matches obtained from the first and second rounds of bipartite matching, respectively.
If $|\mathcal{V}_{j}|=1$, the infrared cluster $\mathbf{m}^{j}_{ir}$ is regarded as
\textbf{\emph{reliable}},
as it corresponds to a unique visible prototype.
Otherwise, it is considered
\textbf{\emph{ambiguous}},
since multiple visible clusters are matched to the same infrared cluster.
Accordingly, all one-to-one correspondences are collected into the reliable set:
\begin{equation}
\mathcal{R}
=
\bigl\{
(\mathbf{m}_{ir}^{j},\,\mathbf{m}_{vis}^{i})
\;\big|\;
|\mathcal{V}_{j}|=1,\ i\in\mathcal{V}_{j}
\bigr\},
\end{equation}
which enumerates every infrared prototype that is uniquely matched to a single visible prototype.
Conversely, all one-to-many relations are grouped into the ambiguous set:
\begin{equation}
\mathcal{A}
=
\bigl\{
(\mathbf{m}_{ir}^{j},\,\{\mathbf{m}_{vis}^{i}\}_{i\in\mathcal{V}_{j}})
\;\big|\;
|\mathcal{V}_{j}|>1
\bigr\}.
\end{equation}

\textbf{Memory Reconstruction.}
Once reliable and ambiguous associations are identified, we reconstruct a 
structurally aligned cross-modality memory bank.
For each reliable pair 
$(\mathbf{m}_{ir}^{j},\mathbf{m}_{vis}^{i})\in\mathcal{R}$,
the prototypes already form a consistent one-to-one correspondence.
Thus, the refined memory simply inherits their original representations:
\begin{equation}
\tilde{\mathbf{m}}_{vis}^{i}=\mathbf{m}_{vis}^{i},
\qquad
\tilde{\mathbf{m}}_{ir}^{i}=\mathbf{m}_{ir}^{j}.
\end{equation}
For each ambiguous group 
$(\mathbf{m}_{ir}^{j},\{\mathbf{m}_{vis}^{i}\}_{i\in\mathcal{V}_{j}})\in\mathcal{A}$,
the infrared cluster must be reorganized to match the finer granularity of its
visible counterparts.
Let $\mathcal{F}_{ir,j}=\{\mathbf{f}_{ir,n}\}_{n\in\mathcal{I}_{j}}$ denote all
infrared features belonging to cluster $j$, where $\mathcal{I}_{j}$ indexes the
samples assigned to this cluster.
Each infrared feature is reassigned to the most similar visible prototype 
indexed by $i^{*}$:
\begin{equation}
i^{*}
=
\arg\max_{i\in\mathcal{V}_{j}}
\operatorname{Sim}(\mathbf{f}_{ir,n},\mathbf{m}_{vis}^{i}),
\end{equation}
Based on these assignments, the coarse infrared cluster is then split into 
mutually aligned sub-clusters:
\begin{equation}
\tilde{\mathcal{F}}_{ir,i}
=
\bigl\{
\mathbf{f}_{ir,n}\in \mathcal{F}_{ir,j}\;\big|\; i^{*}=i
\bigr\},
\quad
i\in\mathcal{V}_{j}.
\end{equation}
The refined infrared prototypes are then computed by aggregating the features within each newly formed sub-cluster:
\begin{equation}
\tilde{\mathbf{m}}_{ir}^{i}
=
\frac{1}{|\tilde{\mathcal{F}}_{ir,i}|}
\sum_{\mathbf{f}_{ir,n}\in\tilde{\mathcal{F}}_{ir,i}}
\mathbf{f}_{ir,n}.
\end{equation}
After refinement, every visible prototype $\mathbf{m}_{vis}^{i}$ is paired with
a unique infrared prototype $\tilde{\mathbf{m}}_{ir}^{i}$, producing a 
structurally consistent memory bank that satisfies:
\begin{equation}
|\tilde{\mathcal{M}}_{vis}| = |\tilde{\mathcal{M}}_{ir}| =  \tilde{K}.
\end{equation}
where $\tilde{K}$ denotes the refined number of visible–infrared prototype
pairs after reconstruction. 

\textbf{Uncertainty Learning.}
Hard reassignment in Memory Reconstruction enforces clear and deterministic cluster boundaries for reliable associations.
However, when an infrared feature exhibits comparable similarity to multiple visible prototypes, ambiguity naturally arises near cluster boundaries.
To prevent error amplification caused by hard decisions, PGUR preserves the soft similarity distribution for such ambiguous features.
For an ambiguous infrared feature 
$\mathbf{f}_{ir,n}\in \tilde{\mathcal{F}}_{ir,j}$ 
and its candidate visible prototype set $\mathcal{V}_{j}$, we compute its soft
target distributions over the visible and infrared prototype banks as:
\begin{equation}
\begin{aligned}
s^{i2v}_{n,i}
&=
\frac{
\exp\!\bigl(\mathbf{f}_{ir,n}\cdot\tilde{\mathbf{m}}^{i}_{vis}\bigr)
}{
\sum_{k\in\mathcal{V}_{j}}
\exp\!\bigl(\mathbf{f}_{ir,n}\cdot\tilde{\mathbf{m}}^{k}_{vis}\bigr)
},
\quad i\in\mathcal{V}_{j},
\\[6pt]
s^{i2i}_{n,i}
&=
\frac{
\exp\!\bigl(\mathbf{f}_{ir,n}\cdot\tilde{\mathbf{m}}^{i}_{ir}\bigr)
}{
\sum_{k\in\mathcal{V}_{j}}
\exp\!\bigl(\mathbf{f}_{ir,n}\cdot\tilde{\mathbf{m}}^{k}_{ir}\bigr)
},
\quad i\in\mathcal{V}_{j}.
\end{aligned}
\label{eq:soft-ir}
\end{equation}
After memory refinement, the two modalities maintain their own aligned
prototype sets  
$\tilde{\mathcal{M}}_{vis}=\{\tilde{\mathbf{m}}^{i}_{vis}\}_{i=1}^{\tilde{K}}$  
and  
$\tilde{\mathcal{M}}_{ir}=\{\tilde{\mathbf{m}}^{i}_{ir}\}_{i=1}^{\tilde{K}}$,  
which share indices but remain modality-specific.  
For an infrared feature $\mathbf{f}_{ir,n}$, its predicted distributions over
the two prototype sets are:
\begin{equation}
\begin{aligned}
p^{i2v}_{n,i}
&=
\frac{
\exp\!\left(\mathbf{f}_{ir,n}\cdot\tilde{\mathbf{m}}^{i}_{vis}/\tau\right)
}{
\sum_{k=1}^{\tilde{K}}
\exp\!\left(\mathbf{f}_{ir,n}\cdot\tilde{\mathbf{m}}^{k}_{vis}/\tau\right)
},
\\[6pt]
p^{i2i}_{n,i}
&=
\frac{
\exp\!\left(\mathbf{f}_{ir,n}\cdot\tilde{\mathbf{m}}^{i}_{ir}/\tau\right)
}{
\sum_{k=1}^{\tilde{K}}
\exp\!\left(\mathbf{f}_{ir,n}\cdot\tilde{\mathbf{m}}^{k}_{ir}/\tau\right)
}.
\end{aligned}
\label{eq:pred-ir}
\end{equation}

To facilitate robust cross-modality alignment while preventing error propagation from ambiguous associations, we formulate the optimization objective by distinguishing between reliable and ambiguous samples.  
The unified objective for the infrared modality is defined as:
\begin{equation}\mathcal{L}_{PGUR}^{ir} = \mathcal{L}_{rel}^{ir} + \lambda_3 \mathcal{L}_{amb}^{ir},\end{equation}
where the hard supervision term $\mathcal{L}_{rel}^{ir}$ for reliable samples is:
\begin{equation}\mathcal{L}_{rel}^{ir} = -\log p^{i2v}_{n, \hat{y}_{ir,n}} - \log p^{i2i}_{n, \hat{y}_{ir,n}},
\end{equation}
where $\hat{y}_{ir,n}$ represents the unique prototype index assigned to sample $\mathbf{f}_{ir,n}$ through the one-to-one matching $(i,j) \in \mathcal{R}$. For ambiguous samples, the soft uncertainty-aware supervision term $\mathcal{L}_{amb}$ is formulated as:
\begin{equation}
    \mathcal{L}_{amb}^{ir} = \sum_{i \in \mathcal{V}_j} \bigl( -s^{i2v}_{n,i} \log p^{i2v}_{n,i} - s^{i2i}_{n,i} \log p^{i2i}_{n,i} \bigr),
\end{equation}
where $\mathcal{V}_j$ is the set of candidate visible prototype indices associated with the cluster of $\mathbf{f}_{ir,n}$, and $s_{n,i}$ are the soft targets defined in Eq. (\ref{eq:soft-ir}).

Symmetrically, the refinement process is applied to the visible modality. By treating visible features $\mathbf{f}_{vis,n}$ as anchors and the aligned prototype sets as targets, the corresponding loss $\mathcal{L}_{PGUR}^{vis}$ is formulated as:
\begin{equation}\mathcal{L}_{PGUR}^{vis} =  \mathcal{L}_{rel}^{vis} + \lambda_4  \mathcal{L}_{amb}^{vis},\end{equation}
The final joint objective for Prototype-Guided Uncertainty Refinement (PGUR) is defined as the summation of the losses from both modalities:
\begin{equation}\mathcal{L}_{PGUR} = \mathcal{L}_{PGUR}^{ir} + \mathcal{L}_{PGUR}^{vis}.
\end{equation}

\subsection{Overall Objective}
The overall objective of the proposed CBA framework integrates intra-modality discriminability, causal representation calibration, and cross-modality structural alignment. The total loss function $\mathcal{L}_{total}$ is formulated as a unified objective:
\begin{equation}\mathcal{L}_{total} = \mathcal{L}_{intra} + \mathcal{L}_{CIW} + \mathcal{L}_{PGUR},
\end{equation}
where $\mathcal{L}_{intra}$ is the intra-modality contrastive loss,
$\mathcal{L}_{CIW} = \mathcal{L}_{mpb} + \lambda_1 \mathcal{L}_{ttb} + \lambda_2 \mathcal{L}_{seq}$
is the causal intervention warm-up loss that purifies the representation space
and is symmetrically applied to both visible and infrared modalities,
and $\mathcal{L}_{PGUR} = \mathcal{L}_{PGUR}^{ir} + \mathcal{L}_{PGUR}^{vis}$
is the prototype-guided uncertainty refinement loss for cross-modal alignment.

\begin{table*}[!ht]\centering\small
\caption{Comparison with state-of-the-art Re-ID methods on HITSZ-VCM in terms of mAP and CMC (\%).
“R@1”, “R@5”, and “R@10” denote Rank-1, Rank-5, and Rank-10 accuracy.
† indicates results from official implementations.
Best and second-best results are highlighted in bold and underlined, respectively.}
\label{vcm}
% 增加了第一列用于显示“Supervised/Unsupervised”，宽度设为很窄 (0.4cm)
\begin{tabular}{m{0.2cm}<{\centering}m{2.3cm}<{\centering}m{2cm}<{\centering}m{1.5cm}<{\centering}m{1cm}<{\centering}m{0.7cm}<{\centering}m{0.7cm}<{\centering}m{0.7cm}<{\centering}m{0.7cm}<{\centering}m{0.7cm}<{\centering}m{0.7cm}<{\centering}m{0.7cm}<{\centering}m{0.7cm}<{\centering}}
\toprule[0.8pt]
\begin{tikzpicture}[overlay, remember picture]
    % 调整了填充范围 (-1.2)，使其覆盖最左侧新增的列
    \fill[gray!30] (-0.34,-0.6) rectangle (17.77,0.31); 
\end{tikzpicture}
% 表头调整，增加了 &
& \multirow{2}*{Methods} & \multirow{2}*{Reference} & \multirow{2}*{Backbone} & \multirow{2}*{Seq\_Len} & \multicolumn{4}{c}{\textit{Infrared-to-Visible}} & \multicolumn{4}{c}{\textit{Visible-to-Infrared}}  
\\ \cmidrule(lr){6-9} \cmidrule(lr){10-13} 
& &  &   &  & {R@1}  & {R@5}  & {R@10}  & mAP    & {R@1}  & {R@5} & {R@10}  & mAP           \\ \toprule[0.8pt]

% --- 监督学习部分 ---
\multirow{8}{*}{\rotatebox{90}{\textbf{Supervised}}} 
& MITML\cite{lin2022learning}      &CVPR'22  &ResNet-50 &6    &63.7 &76.9 &81.7  &45.3 &64.5 &79.0 &83.0 &47.7   \\
& IBAN\cite{li2023intermediary}    &TCSVT'23 &ResNet-50 &6    &65.0 &78.3 &83.0  &48.8 &69.6 &81.5 &85.4 &51.0   \\
% & SADSTRM\cite{li2023adversarial}  &Arxiv'23 &ResNet-50 &6    &65.3 &77.9 &82.7  &49.5 &67.7 &80.7 &85.1 &51.8   \\
& SAADG\cite{zhou2023video}        &ACM MM'23 &ResNet-50 &6    &69.2 &80.6 &85.0  &53.8 &73.1 &83.5 &86.9 &56.1   \\
& CST\cite{feng2024cross}          &TMM'24  &ResNet-50  &6    &69.4 &81.1 &85.8  &51.2 &72.6 &83.4 &86.7 &53.0   \\
& AuxNet\cite{du2023video}         &TIFS'24 &ResNet-50  &6     &51.1 &- &-  &46.0 &54.6 &- &- &48.7   \\
& FA-Net\cite{yang2025fanet}     &TIP'25   &ResNet-50   &6    & 68.1 & 79.7 & 84.0 & 51.2  & 70.0 & 82.1 & 86.0 & 52.5 \\
& VLD\cite{li2025video}     &TIFS'25                       &ViT-B/16   &6    & 74.3 & 85.0 & 88.4 & 60.2  & 74.6 & 86.4 & 90.0 & 58.6 \\
 \hline

% --- 无监督学习部分 (包含 ResNet 和 ViT) ---
\multirow{9}{*}{\rotatebox{90}{\textbf{Unsupervised}}}
& ADCA$^{\dagger}$ \cite{yang2022augmented}   &ACM MM'22 &ResNet-50 &6  &33.1 &45.4 &53.1 &23.2    &35.7 &52.4 &60.0 &23.9  \\
& PGM$^{\dagger}$ \cite{wu2023unsupervised}   &CVPR'23 &ResNet-50 &6 &33.7 &47.1 &53.6 &23.4     &36.9 &52.7 &60.4 &23.9 \\
& GUR$^{\dagger}$ \cite{yang2023towards}   &ICCV'23 &ResNet-50 &6 &9.0 &18.6 &24.7 &5.6      &9.9 &20.1 &26.2 &5.9 \\
& NG$^{\dagger}$ \cite{cheng2023unsupervised}    &ACM MM'23 &ResNet-50 &6 &22.8 &38.2 &47.0 &12.0     &24.6 &40.2 &48.5 &12.1 \\
\cline{2-13} % 只在右侧画横线，不穿过最左侧的标签列
& ADCA$^{\dagger}$ \cite{yang2022augmented}   &ACM MM'22 &ViT-B/16 &6  &30.0 &43.4 &51.7 &24.8    &34.0 &52.3 &61.0 &24.8  \\
&  PGM$^{\dagger}$ \cite{wu2023unsupervised}   &CVPR'23 &ViT-B/16 &6  &\underline{38.4} &\underline{54.6}  &\underline{62.9} &\underline{30.6}    &\underline{40.4} &\underline{59.4} &\underline{68.6} &\underline{29.0} \\
&  RPNR$^{\dagger}$ \cite{yin2024robust}   &ACM MM'24 &ViT-B/16 &6  &28.2 &42.3 &50.2 &22.1    &33.3 &50.9 &59.4 &22.3 \\
& N-ULC$^{\dagger}$\cite{teng2025relieving} &AAAI'25 &ViT-B/16 &6  &28.2 &42.3 &50.2 &22.1    &33.3 &50.9 &59.4 &22.3 \\
\cline{2-13}
 \rowcolor{gray!10} & our  & -  & ViT-B/16  & 6 & \textbf{58.6} & \textbf{74.9} & \textbf{80.7} & \textbf{45.9} & \textbf{63.0} & \textbf{77.8} & \textbf{83.4} & \textbf{45.1} \\
 \toprule[0.8pt]

\end{tabular}
\end{table*}
\begin{table*}[!ht]\centering\small
\caption{Comparison with state-of-the-art Re-ID methods on BUPTCampus in terms of mAP and CMC (\%).
† indicates results from official implementations.
Best and second-best results are highlighted in bold and underlined, respectively.}
\label{bupt}
% 增加了第一列用于显示“Supervised/Unsupervised”，并调整了后面列的对齐
\begin{tabular}{m{0.2cm}<{\centering}m{2.3cm}<{\centering}m{2cm}<{\centering}m{1.5cm}<{\centering}m{1cm}<{\centering}m{0.7cm}<{\centering}m{0.7cm}<{\centering}m{0.7cm}<{\centering}m{0.7cm}<{\centering}m{0.7cm}<{\centering}m{0.7cm}<{\centering}m{0.7cm}<{\centering}m{0.7cm}<{\centering}}
\toprule[0.8pt]
\begin{tikzpicture}[overlay, remember picture]
    % 调整灰色标题栏填充范围，使其覆盖新增的最左侧列
    \fill[gray!30] (-0.34,-0.6) rectangle (17.77,0.31); 
\end{tikzpicture}
% 表头调整
& \multirow{2}*{Methods} & \multirow{2}*{Reference} & \multirow{2}*{Backbone} & \multirow{2}*{Seq\_Len} & \multicolumn{4}{c}{\textit{Infrared-to-Visible}} & \multicolumn{4}{c}{\textit{Visible-to-Infrared}} \\ 
\cmidrule(lr){6-9} \cmidrule(lr){10-13} 
& & & & & {R@1} & {R@5} & {R@10} & mAP & {R@1} & {R@5} & {R@10} & mAP \\ \toprule[0.8pt]

% --- 有监督部分 ---
\multirow{3}*{\rotatebox{90}{\textbf{Supervised}}} 
& MITML\cite{lin2022learning}    &CVPR'22  &ResNet-50 &6  &49.1 &67.9 &75.4 &47.5 &50.2 &68.3 &75.7 &46.3   \\
& AuxNet\cite{du2023video}       &TIFS'24  &ResNet-50 &10 &63.6 &79.9 &85.3 &61.1 &62.7 &81.5 &85.7 &60.2    \\

& TF-CLIP \cite{li2025video}    &AAAI'25              &ViT-B/16  
&6   
&57.1 &80.4 &89.4 &56.6 
&59.6 &78.7 &87.4 &55.3   \\
& VLD \cite{li2025video}    &TIFS'25              &ViT-B/16  &6   &65.3 &84.9 &89.7 &63.5 &65.8 &83.0 &87.9 &63.0   \\
& X-ReID \cite{li2025video}    &AAAI'26              &ViT-B/16  &6   
&68.2 &88.4 &94.3 &68.5 
&68.8 &84.8 &92.7 &65.9   \\
 \hline

% --- 无监督部分 ---
\multirow{9}*{\rotatebox{90}{\textbf{Unsupervised}}}
& ADCA$^{\dagger}$ \cite{yang2022augmented}   &ACM MM'22 &ResNet-50 &6 &27.6 &47.0 &55.4 &27.8 &29.6 &49.3 &58.2 &29.1 \\
& PGM$^{\dagger}$ \cite{wu2023unsupervised}   &CVPR'23 &ResNet-50 &6 &\underline{29.5} &48.1 &56.7 &30.1 &32.2 &51.1 &57.8 &30.9 \\
& GUR$^{\dagger}$ \cite{yang2023towards}   &ICCV'23 &ResNet-50 &6 &12.9 &25.2 &33.0 &14.4 &17.0 &29.1 &36.0 &17.7 \\
& NG$^{\dagger}$ \cite{cheng2023unsupervised}    &ACM MM'23 &ResNet-50 &6 &26.1 &41.2 &46.1 &25.4 &27.4 &41.3 &48.7 &25.0 \\
\cline{2-13}
& ADCA$^{\dagger}$ \cite{yang2022augmented}   &ACM MM'22 &ViT-B/16 &6 &18.8 &38.1 &48.1 &21.8 &25.8 &43.4 &51.2 &26.0 \\
& PGM$^{\dagger}$ \cite{wu2023unsupervised}   &CVPR'23 &ViT-B/16 &6 &27.0 &\underline{51.3} &\underline{60.5} &\underline{28.9} &\underline{33.6} &\underline{55.1} &\underline{62.3} &\underline{33.4} \\
& RPNR$^{\dagger}$ \cite{yin2024robust}   &ACM MM'24 &ViT-B/16 &6 &24.0 &42.4 &51.4 &24.6 &20.3 &38.5 &46.2 &22.4 \\
& N-ULC$^{\dagger}$ \cite{teng2025relieving} &AAAI'25 &ViT-B/16 &6 &22.8 &45.2 &52.1 &25.2 &29.1 &49.0 &59.0 &28.8 \\
\cline{2-13}
 \rowcolor{gray!10} & our      &-                        &ViT-B/16   &6    & \textbf{37.7} & \textbf{58.8} & \textbf{67.6} & \textbf{37.1}  & 
 \textbf{40.2} & \textbf{61.5} & \textbf{69.0} & \textbf{38.8} \\
 \toprule[0.8pt]
\end{tabular}
\end{table*}

\section{Experiments}
\subsection{Datasets and Evaluation Metrics}

\noindent\textbf{Datasets.}
HITSZ-VCM~\cite{lin2022learning} is collected using paired visible and infrared cameras,
containing 927 identities captured by 6 visible and 6 infrared cameras.
It includes 251,452 RGB frames and 211,807 infrared frames, with an average of 24 frames per tracklet.
Following the standard protocol, 500 identities are used for training and 427 for testing.
BUPTCampus~\cite{du2023video} is a large-scale dataset collected by 6 cameras,
comprising 3,080 identities and 16,826 tracklets with 1.87 million frames.
It is split into three subsets: 1,074 identities for primary training,
930 for auxiliary supervision, and 1,076 for evaluation.

\noindent\textbf{Evaluation Metrics.}
Retrieval performance is evaluated using CMC and mAP.
Following prior work~\cite{lin2022learning,du2023video},
results are reported under infrared-to-visible and visible-to-infrared settings.

\vspace{-6.5pt}
\subsection{Implementation Details}
The proposed CBA framework is implemented in PyTorch and evaluated on a single
NVIDIA RTX 5090 GPU. 
% For reproducibility, random seeds for Python, NumPy, and
% PyTorch are fixed to 1. 
The visual encoder follows the VLD framework with
spatio-temporal prompting (STP)~\cite{li2025video}, built on CLIP ViT-B/16. STP
explicitly models temporal continuity across frames. Input images are resized to
$288\times144$ and augmented with random flipping, padding, cropping, and channel
erasure~\cite{ye2021channel}.
We adopt the Adam optimizer and train the model in three stages. The first stage
runs for 40 epochs with a learning rate of $4\times10^{-5}$ and cosine decay,
activating CIW with a batch size of 16. The second
stage performs intra-modality contrastive learning for 30 epochs with the
learning rate reduced by $100\times$. The third stage enables
PGUR and continues training for
50 epochs with the same reduced learning rate.
Unless otherwise specified, the batch size is set to 32, with 16 sequences per
modality per mini-batch and 6 frames sampled from each sequence. The hyperparameters $\lambda_1$, $\lambda_2$, $\lambda_3$, and $\lambda_4$
are set to 2.6, 0.1, 1.9, and 1.5, respectively.

% \vspace{-4pt}
\subsection{Comparison with State-of-the-Art Methods}
In this section, we compare our method with state-of-the-art approaches on
HITSZ-VCM and BUPTCampus.
Since most unsupervised cross-modality Re-ID methods are image-based, we evaluate
representative methods (e.g., ADCA~\cite{yang2022augmented}, PGM~\cite{wu2023unsupervised},
GUR~\cite{yang2023towards}, NG~\cite{cheng2023unsupervised}, RPNR~\cite{yin2024robust},
and N-ULC~\cite{teng2025relieving}) on the two video benchmarks using their
official implementations.
All methods are trained following their official implementations, with only
necessary adaptations for video inputs and the corresponding evaluation
protocols.
To ensure architectural fairness, methods supporting ViT-B/16 are further
evaluated with the same backbone.

\textbf{Evaluation on the HITSZ-VCM Dataset.}
We evaluate our method on HITSZ-VCM with Seq\_Len=6 under both
Infrared-to-Visible (I2V) and Visible-to-Infrared (V2I) settings, as shown in
Table~\ref{vcm}.
Focusing on USL-VVI-ReID, we
mainly compare with representative USL-VI-ReID methods.
Our approach achieves the best performance among all unsupervised competitors in
both directions.
Under the ViT-B/16 setting, it attains 58.6\% Rank-1 and 45.9\% mAP in I2V, and
63.0\% Rank-1 and 45.1\% mAP in V2I.
Compared with the strongest unsupervised baseline under the same backbone (PGM),
our method improves Rank-1/mAP by +20.2\%/+15.3\% in I2V and
+22.6\%/+16.1\% in V2I, demonstrating substantially enhanced cross-modality
matching on video sequences.
These gains mainly arise from the proposed CBA framework, where CIW produces more
reliable identity representations for clustering, and PGUR further corrects
granularity mismatch via prototype-guided refinement with uncertainty-aware
supervision.

\textbf{Evaluation on the BUPTCampus Dataset.}
BUPTCampus is a more challenging benchmark with larger scale and higher
cross-modality variability.
As shown in Table~\ref{bupt}, our method achieves the best
performance in both I2V and V2I
settings.
Specifically, CBA attains 37.7\% Rank-1 and 37.1\% mAP in I2V, and
40.2\% Rank-1 and 38.8\% mAP in V2I.
Compared with the strongest unsupervised baseline under the same ViT-B/16
backbone (PGM), our method improves Rank-1/mAP by +10.7\%/+8.2\% in I2V and
+6.6\%/+5.4\% in V2I.
These results demonstrate that CIW and PGUR remain effective under larger-scale
data and stronger modality discrepancies, validating the robustness and
scalability of the proposed framework.

\begin{table*}[!ht]\small
 \centering {\caption{Ablation study of CBA components on HITSZ-VCM.
Starting from a pre-trained encoder, we progressively add the baseline pipeline (B), CIW, and PGUR.
Performance is reported for Infrared-to-Visible and Visible-to-Infrared retrieval.
Numbers in parentheses indicate gains over the pre-trained baseline.}\label{component}
\begin{tikzpicture}[overlay, remember picture]
    \fill[gray!30] (0.1,0.8) rectangle (15.7,1.72); % adjust coordinates to match your table cells
\end{tikzpicture}
% \begin{tikzpicture}[overlay, remember picture]
%     \fill[gray!10] (0.11,-0.77) rectangle (13.95,-1.1); % adjust coordinates to match your table cells
% \end{tikzpicture}
\begin{tabular}{m{1.3cm}<{\centering}m{1.1cm}<{\centering}m{1.3cm}<{\centering}m{1.3cm}<{\centering}m{1.8cm}<{\centering}m{1.8cm}<{\centering}m{1.8cm}<{\centering}m{1.8cm}<{\centering}}
\toprule[0.8pt] 
 \multirow{2}{*}{Pre-trained}  &\multicolumn{3}{c}{Component}  & \multicolumn{2}{c}{\textit{Infrared-to-Visible}}   & \multicolumn{2}{c}{\textit{Visible-to-Infrared}} \\ 
% \cline{2-5} 
\cmidrule(lr){2-4} \cmidrule(lr){5-6} \cmidrule(lr){7-8} 
% \cline{6-7} \cline{8-9}
 & B &CIW & PGUR      & R@1    & mAP & R@1    & mAP \\ 
\toprule[0.8pt]
\cmark &\xmark &\xmark   &\xmark           
& 1.1  & 1.1  & 4.1  & 2.3 \\
\cmark &\xmark &\cmark   &\xmark           
& 18.9\gtext{(+17.8)}  & 12.3\gtext{(+11.2)}  &18.2\gtext{(+14.1)}  &11.3\gtext{(+9.0)}\\
\hline
\cmark &\cmark &\xmark   &\xmark           
& 26.9\gtext{(+25.8)}  & 21.7\gtext{(+20.6)}  &27.9\gtext{(+23.8)}  &20.6\gtext{(+18.3)}\\
 \cmark &\cmark &\cmark   &\xmark          
 & 36.3\gtext{(+35.2)}  & 29.0\gtext{(+27.9)}  & 39.6\gtext{(+35.5)}  & 28.8\gtext{(+26.5)} \\
 \cmark &\cmark &\xmark   &\cmark       
 & 47.8\gtext{(+46.7)}  & 35.1\gtext{(+34.0)}  & 48.8\gtext{(+44.7)}  & 33.3\gtext{(+31.0)} \\
  \rowcolor{gray!10}
\cmark & \cmark &\cmark   &\cmark      
& 58.6\gtext{(+57.5)}  & 45.9\gtext{(+44.8)}   & 63.0\gtext{(+58.9)}  & 45.1\gtext{(+42.8)} \\
\toprule[0.8pt]
\end{tabular}}
\end{table*}

\begin{table*}[!ht]\small
 \centering {\caption{Detailed ablation study on the CIW constituents. The results demonstrate the progressive improvements achieved by MPB, TTB, and ICS across both infrared-to-visible and visible-to-infrared evaluation protocols.}\label{CIW}
\begin{tikzpicture}[overlay, remember picture]
    \fill[gray!30] (0.1,0.44) rectangle (15.7,1.36); % adjust coordinates to match your table cells
\end{tikzpicture}
\begin{tabular}{m{1.3cm}<{\centering}m{1.1cm}<{\centering}m{1.3cm}<{\centering}m{1.3cm}<{\centering}m{1.8cm}<{\centering}m{1.8cm}<{\centering}m{1.8cm}<{\centering}m{1.8cm}<{\centering}}
\toprule[0.8pt] 
 \multirow{2}{*}{B\&PGUR}  &\multicolumn{3}{c}{CIW}  & \multicolumn{2}{c}{\textit{Infrared-to-Visible}}   & \multicolumn{2}{c}{\textit{Visible-to-Infrared}} \\ 
% \cline{2-5} 
\cmidrule(lr){2-4} \cmidrule(lr){5-6} \cmidrule(lr){7-8} 
% \cline{6-7} \cline{8-9}
 & MPB &TTB & ICS      & R@1    & mAP & R@1    & mAP \\ 
\toprule[0.8pt]
\cmark &\xmark &\xmark   &\xmark           
 & 47.8  & 35.1  & 48.8  & 33.3 \\
\cmark &\cmark &\xmark   &\xmark           
& 52.9\gtext{(+5.1)}  & 40.2\gtext{(+5.1)}  &55.0\gtext{(+6.2)}  &37.5\gtext{(+4.2)}\\
%  \cmark &\xmark &\cmark   &\xmark          
%  & 34.6\gtext{(+34.8)}  & 25.9\gtext{(+29.2)}  & 41.0\gtext{(+36.4)}  & 27.3\gtext{(+27.9)} \\
%  \cmark &\xmark &\xmark   &\cmark       
%  & 52.2\gtext{(+46.7)}  & 40.6\gtext{(+34.0)}  & 54.3\gtext{(+44.7)}  & 40.1\gtext{(+31.0)} \\
  \cmark &\cmark &\cmark   &\xmark       
 & 55.8\gtext{(+8.0)}  & 42.2\gtext{(+7.1)}  & 57.1\gtext{(+8.3)}  & 38.8\gtext{(+5.5)} \\
  \rowcolor{gray!10}
\cmark & \cmark &\cmark   &\cmark      
& 58.6\gtext{(+10.8)}  & 45.9\gtext{(+10.8)}   & 63.0\gtext{(+14.2)}  & 45.1\gtext{(+11.8)} \\
\toprule[0.8pt]
\end{tabular}}
\end{table*}

\begin{table*}[!ht]\small
 \centering 
\caption{Ablation of style intervention (SI) in CIW under the CBA framework.
We progressively enable SI in one direction (V$\rightarrow$I or I$\rightarrow$V) or both directions.
}
 \label{mpb}\begin{tikzpicture}[overlay, remember picture]
    \fill[gray!30] (0.1,0.44) rectangle (13.37,1.36); % adjust coordinates to match your table cells
\end{tikzpicture}
\begin{tabular}{m{2.3cm}<{\centering}m{1.4cm}<{\centering}m{1.4cm}<{\centering}m{1.3cm}<{\centering}m{1.3cm}<{\centering}m{1.3cm}<{\centering}m{1.3cm}<{\centering}}
\toprule[0.8pt]
 \multirow{2}{*}{CBA (w/o MPB)} & \multicolumn{2}{c}{MPB} & \multicolumn{2}{c}{\textit{Infrared-to-Visible}} & \multicolumn{2}{c}{\textit{Visible-to-Infrared}} \\ 
 \cmidrule(lr){2-3} \cmidrule(lr){4-5} \cmidrule(lr){6-7} 
  & $V \to I$ & $I \to V$ & R@1 & mAP & R@1 & mAP \\ 
\toprule[0.8pt]
 \cmark     & \xmark & \xmark 
 & 55.5 & 43.2 & 57.1 & 41.2 \\
 \cmark   & \cmark & \xmark 
 & 56.5\gtext{(+1.0)} & 42.2\rtext{(-1.0)} & 59.2\gtext{(+2.1)} & 40.9\rtext{(-0.3)} \\
 \cmark   & \xmark & \cmark 
 & 55.4\rtext{(-0.1)} & 41.9\rtext{(-1.3)} & 58.0\gtext{(+0.9)} & 40.2\rtext{(-1.0)} \\
 \rowcolor{gray!10}
 \cmark   & \cmark & \cmark 
& 58.6\gtext{(+3.1)}  & 45.9\gtext{(+2.7)}   & 63.0\gtext{(+5.9)}  & 45.1\gtext{(+3.9)} \\
\toprule[0.8pt]
\end{tabular}
\end{table*}

\begin{table*}[!ht]\small
 \centering 
\caption{Ablation of PGUR supervision on reliable and ambiguous samples.
We progressively enable hard supervision for reliable pairs and uncertainty-aware soft supervision for ambiguous ones.}
 \label{table_pgur_ablation}
 \begin{tikzpicture}[overlay, remember picture]
    \fill[gray!30] (0.1,0.27) rectangle (13.36,1.17); % adjust coordinates to match your table cells
\end{tikzpicture}
\begin{tabular}{m{2.3cm}<{\centering}m{1.4cm}<{\centering}m{1.4cm}<{\centering}m{1.3cm}<{\centering}m{1.3cm}<{\centering}m{1.3cm}<{\centering}m{1.3cm}<{\centering}m{1.3cm}<{\centering}}
\toprule[0.8pt]
 \multirow{2}{*}{CBA (w/o PGUR) }  & \multicolumn{2}{c}{PGUR} & \multicolumn{2}{c}{\textit{Infrared-to-Visible}} & \multicolumn{2}{c}{\textit{Visible-to-Infrared}} \\ 
 \cmidrule(lr){2-3} \cmidrule(lr){4-5} \cmidrule(lr){6-7} 
  & Reliable & Ambiguous & R@1 & mAP & R@1 & mAP \\ 
\toprule[0.8pt]
 \cmark     & \xmark & \xmark 
 & 36.3  & 29.0  & 39.6  & 28.8 \\
 \cmark   & \cmark & \xmark 
 & 49.2\gtext{(+12.9)} & 39.0\gtext{(+10.0)} & 53.0\gtext{(+13.4)} & 39.3\gtext{(+10.5)} \\
 \rowcolor{gray!10}
 \cmark   & \cmark & \cmark 
& 58.6\gtext{(+22.3)}  & 45.9\gtext{(+16.9)}   & 63.0\gtext{(+23.4)}  & 45.1\gtext{(+16.3)}  \\
\toprule[0.8pt]
\end{tabular}
\end{table*}

% \vspace{-4pt}
\subsection{Ablation Study}
To evaluate individual components of CBA, we conduct ablation studies on
HITSZ-VCM.
Following ADCA~\cite{yang2022augmented}, we adopt as the baseline a standard
unsupervised pipeline that alternates intra-modality clustering and
contrastive learning, without CIW and PGUR.
Relying solely on modality-specific pseudo labels, this baseline is prone to
learning modality-dependent representations.

\begin{figure}[t!]
\centering
\includegraphics[width=8.8cm,keepaspectratio=true]{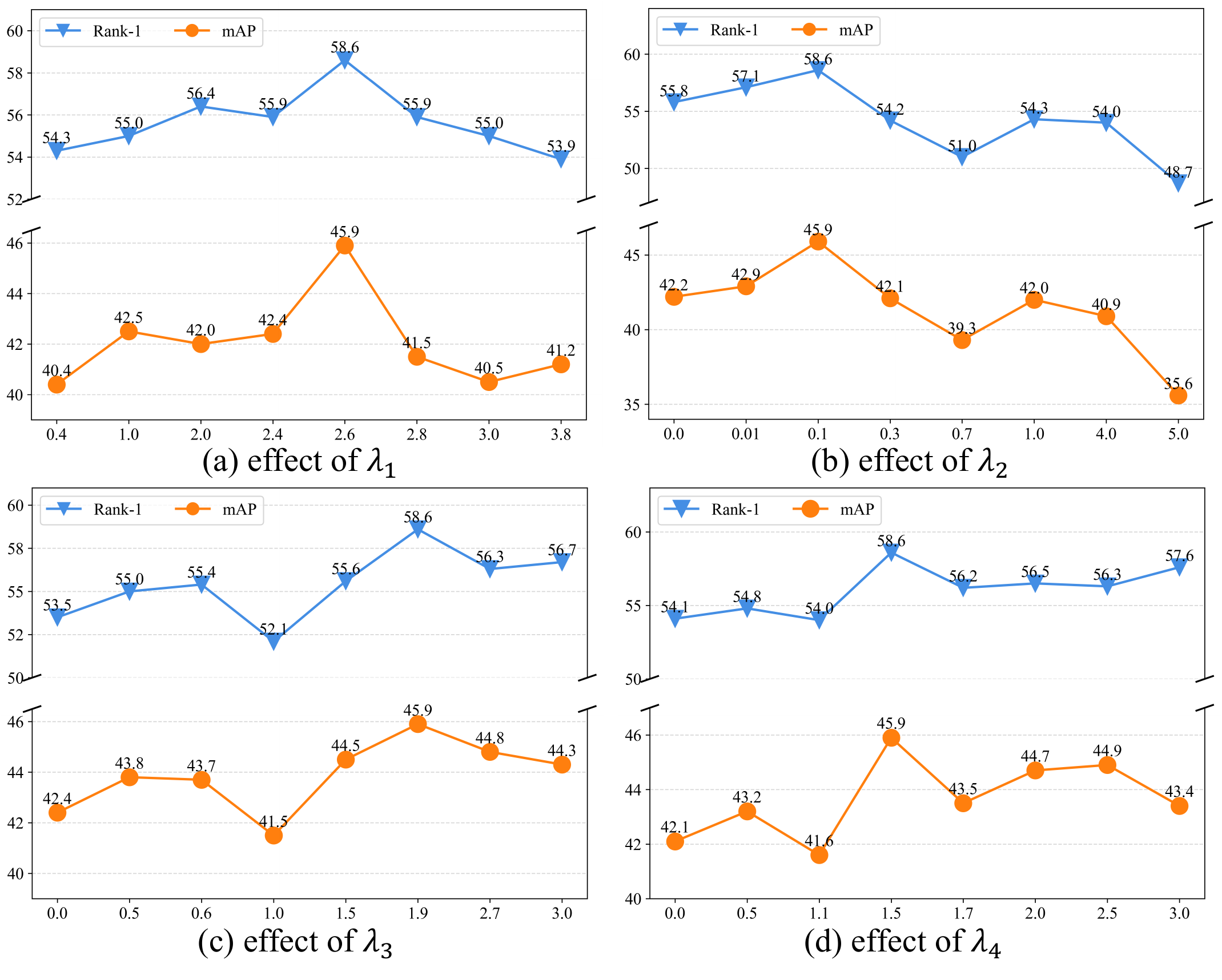}
\caption{Results of Rank-1 and mAP under different values of $\lambda_1$, $\lambda_2$, $\lambda_3$ and $\lambda_4$ on the HITSZ-VCM dataset.}
\label{fig:param}
\end{figure}
\begin{figure}[t!]
\centering
\includegraphics[width=8.8cm,keepaspectratio=true]{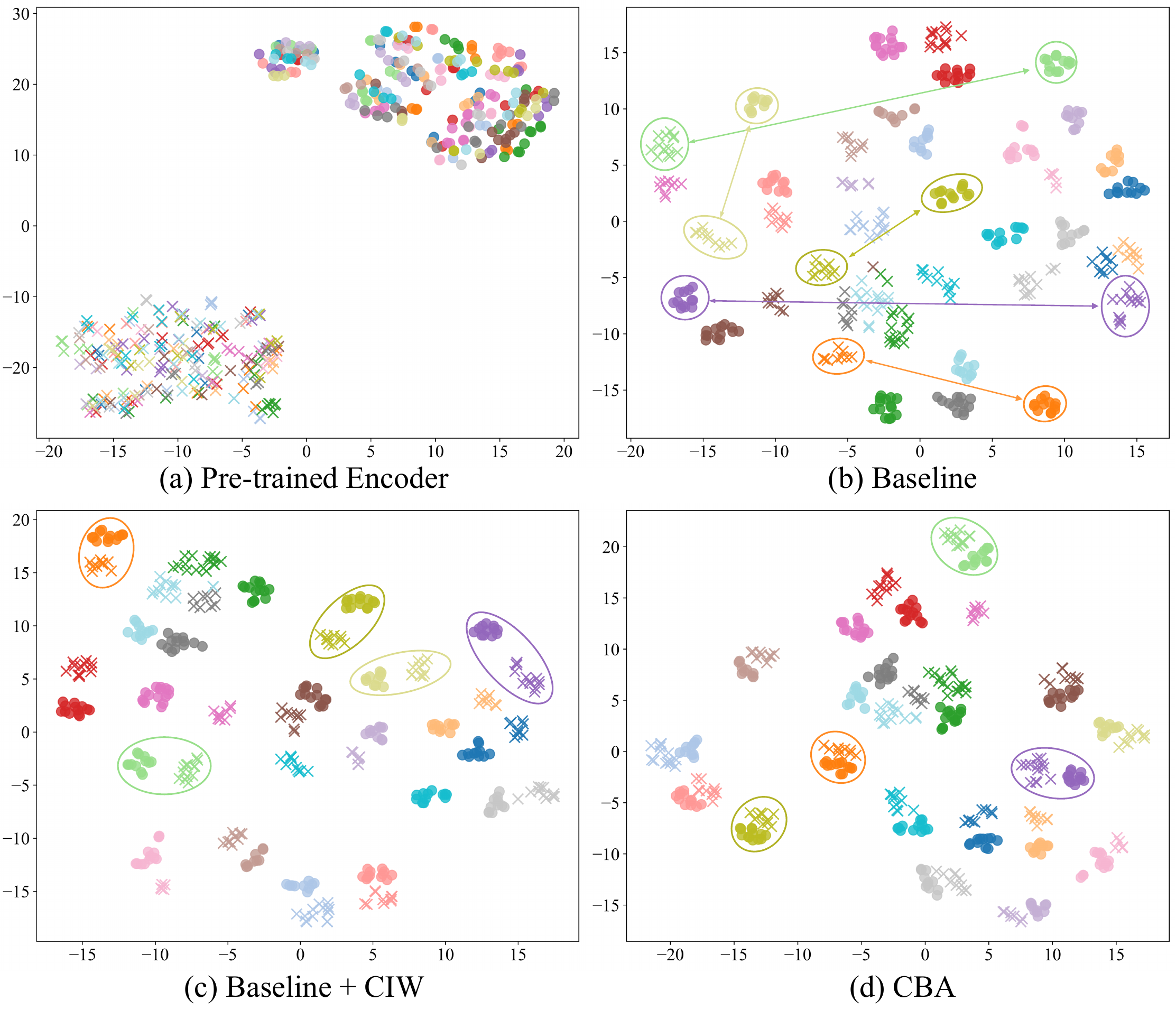}
\caption{
t-SNE visualization of cross-modality embeddings on HITSZ-VCM.
Circles and crosses denote visible and infrared modalities, respectively, while colors indicate identities.
}
\label{tsne}
\end{figure}

\begin{figure}[t!]
\centering
\includegraphics[width=8.8cm,keepaspectratio=true]{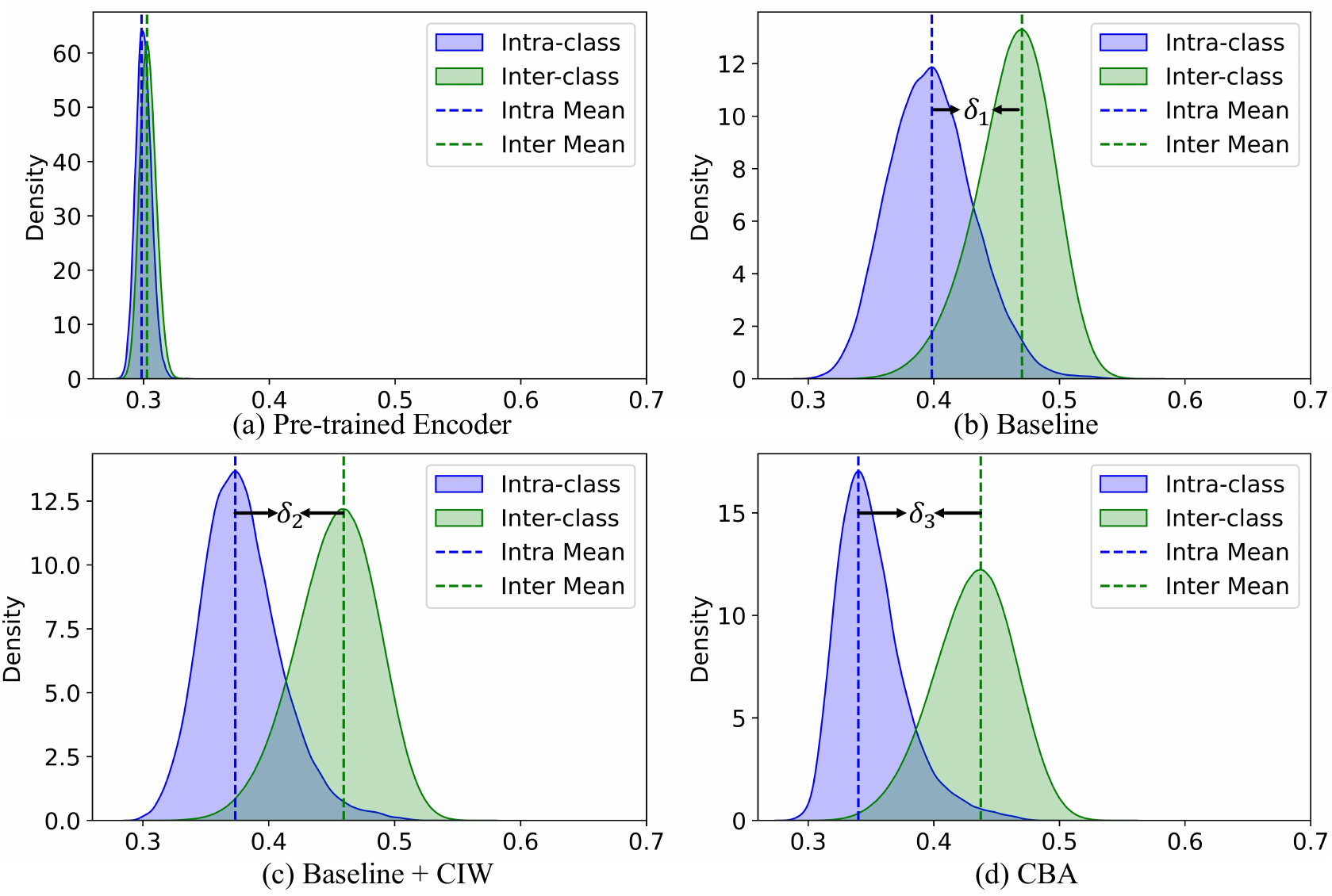}
\caption{
Distributions of intra- and inter-class distances on HITSZ-VCM.
$\delta_1$, $\delta_2$, and $\delta_3$ denote the separation margins between the
intra- and inter-class distributions under different settings, where a larger
margin indicates better identity discriminability.
}
\label{dis}
\end{figure}

\textbf{\textit{1) Analysis of CIW.}}
We evaluate CIW from both holistic and component-wise perspectives, analyzing its overall effectiveness and the complementary roles of MPB, TTB, and ICS (Table~\ref{component}, Table~\ref{CIW} and Table~\ref{mpb}).

\textbf{Overall effectiveness of CIW.}
Table~\ref{component} analyzes the impact of CIW under different configurations.
Starting from the pre-trained encoder, enabling CIW alone yields substantial
performance gains, improving Rank-1/mAP from 1.1\%/1.1\% to 18.9\%/12.3\% in I2V and from 4.1\%/2.3\% to 18.2\%/11.3\% in V2I.
When integrated into the baseline (B), CIW consistently enhances
performance, boosting Rank-1/mAP from 26.9\%/21.7\% to 36.3\%/29.0\% in I2V and from 27.9\%/20.6\% to 39.6\%/28.8\% in V2I.
CIW further complements PGUR. Adding CIW to ``B+PGUR'' improves Rank-1/mAP from 47.8\%/35.1\% to 58.6\%/45.9\% in I2V and from 48.8\%/33.3\% to 63.0\%/45.1\% in V2I. Overall, CIW effectively shifts the representation space from modality-driven to identity-aware, providing a stronger foundation for both intra-modality pseudo-label learning and cross-modality refinement.

\begin{figure*}[t!]
\centering
\includegraphics[width=17cm,keepaspectratio=true]{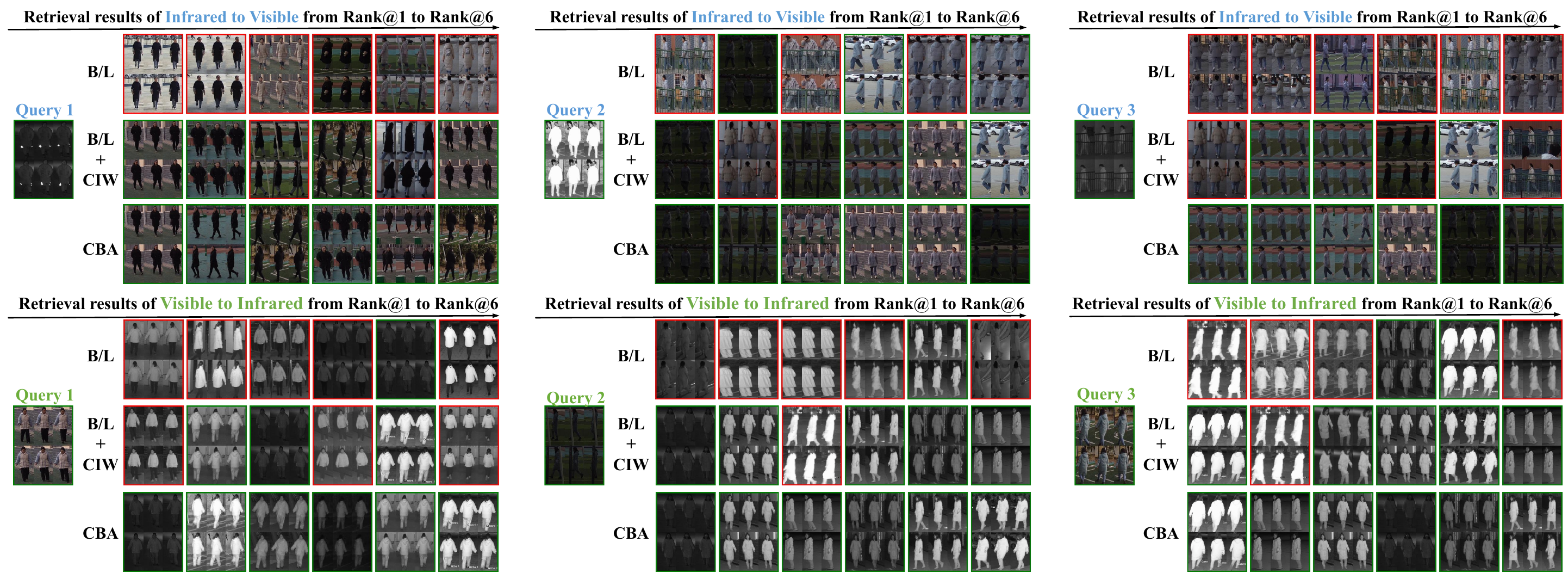}
\caption{Qualitative cross-modality retrieval results on HITSZ-VCM.
Top-6 matches (Rank@1–Rank@6) are shown for Infrared-to-Visible (top) and Visible-to-Infrared (bottom) queries.
Green and red boxes indicate correct and incorrect matches, respectively.}
\label{retrieval}
\end{figure*}

\textbf{Effect of Individual Causal Interventions in CIW.}
Table~\ref{CIW} reports the internal ablation of CIW by progressively activating MPB, TTB, and ICS on top of the baseline with PGUR. Without CIW, the model achieves 47.8\% Rank-1 / 35.1\% mAP (I2V) and 48.8\% / 33.3\% (V2I), indicating residual modality and temporal bias. Introducing MPB yields clear gains (+5.1\% Rank-1, +5.1\% mAP in I2V; +6.2\% Rank-1, +4.2\% mAP in V2I), confirming that modality-specific appearance is a major confounder. Adding TTB further improves robustness to temporal shortcuts, raising performance to 55.8\% / 42.2\% (I2V) and 57.1\% / 38.8\% (V2I). Finally, incorporating ICS delivers the best results, reaching 58.6\% Rank-1 / 45.9\% mAP (I2V) and 63.0\% / 45.1\% (V2I), with total gains of +10.8\% mAP (I2V) and +11.8\% mAP (V2I) over the CIW-free baseline. These results demonstrate that CIW benefits from the complementary effects of modality perturbation, temporal intervention, and identity-consistency stabilization, jointly suppressing non-causal shortcuts while preserving identity-discriminative information.

\textbf{Effectiveness of Bidirectional Intervention in MPB.}
Table~\ref{mpb} evaluates the bidirectional intervention strategy in MPB.
% When only unidirectional intervention is applied ($V\!\to\!I$ or $I\!\to\!V$),
% performance degrades relative to the baseline (CBA w/o MPB), with mAP dropping by
% $1.3\%$ in I2V and $0.3\%$ in V2I.
When only a single-direction intervention is applied,
performance degrades relative to the baseline (CBA w/o MPB).
Specifically, applying only $I\!\to\!V$ leads to an I2V mAP drop of $1.3\%$,
while applying only $V\!\to\!I$ results in a V2I mAP drop of $0.3\%$. 
This degradation stems from asymmetric distribution shifts, where perturbing a
single modality introduces unilateral semantic noise and enlarges cross-modality
discrepancy.
In contrast, enabling bidirectional intervention yields clear gains, improving
mAP by $2.7\%$ in I2V and $3.9\%$ in V2I.
These results indicate that MPB must be applied symmetrically: bidirectional style
perturbation effectively neutralizes modality-specific appearance confounders $M$,
thereby enforcing the separation property ($S \perp M$) and allowing the encoder to
focus on intrinsic identity-related causal factors $S$ for robust cross-modality
retrieval.

\textbf{\textit{2) Analysis of PGUR.}}
PGUR is evaluated by analyzing its overall effectiveness across different pipelines and the complementary roles of hard supervision on reliable pairs and soft supervision on ambiguous samples (Table~\ref{component} and Table~\ref{table_pgur_ablation}).

\textbf{Overall effectiveness of PGUR.}
We evaluate PGUR by integrating it into different pipelines in
Table~\ref{component}. When added to the baseline B, PGUR brings
substantial improvements, boosting Rank-1/mAP from 26.9\%/21.7\% to
47.8\%/35.1\% in I2V and from 27.9\%/20.6\% to 48.8\%/33.3\% in V2I.
Notably, PGUR is more effective when applied after CIW.
With higher-quality pseudo labels and a more reliable clustering
structure, adding PGUR to ``B+CIW'' further improves Rank-1/mAP to
58.6\%/45.9\% in I2V and 63.0\%/45.1\% in V2I.
These results show that PGUR provides effective cross-modality
refinement by handling ambiguous associations with uncertainty-aware
soft supervision, leading to more modality-invariant representations
and improved retrieval performance.

\textbf{Contributions of Reliable and Ambiguous Samples.}
We further analyze PGUR by progressively enabling its supervision on
reliable samples and then on ambiguous ones, as reported in
Table~\ref{table_pgur_ablation}. Without PGUR, CBA achieves
36.3\%/29.0\% Rank-1/mAP in I2V and 39.6\%/28.8\% in V2I.
Enabling PGUR only on reliable associations substantially improves
performance to 49.2\%/39.0\% in I2V and 53.0\%/39.3\% in V2I.
When ambiguous samples are further incorporated with uncertainty-aware
supervision, performance increases to 58.6\%/45.9\% in I2V and
63.0\%/45.1\% in V2I.
These results demonstrate that explicitly modeling ambiguity is crucial:
while reliable pairs benefit from hard supervision, ambiguous samples
require soft, uncertainty-aware constraints to avoid error propagation
and achieve robust cross-modality alignment.

\subsection{Parameter Analysis}
We analyze the sensitivity of the hyper-parameters $\lambda_1$, $\lambda_2$, $\lambda_3$, and $\lambda_4$ under the I2V protocol on HITSZ-VCM.
Here, $\lambda_1$ and $\lambda_2$ control the strength of temporal intervention and identity-consistency stabilization in CIW, while $\lambda_3$ and $\lambda_4$ weight the same form of uncertainty-aware soft alignment loss in PGUR, applied to the infrared and visible modalities, respectively.
As shown in Fig.~\ref{fig:param}, all parameters exhibit a similar trend:
performance improves as the corresponding constraint becomes effective, and degrades when over-weighted due to over-regularization or over-alignment.
We therefore set $\lambda_1 = 2.6$, $\lambda_2 = 0.1$, $\lambda_3 = 1.9$, and $\lambda_4 = 1.5$ in all experiments.

\subsection{Visualization}
We provide qualitative analysis via t-SNE visualization, distance distribution
analysis, and cross-modality retrieval examples under I2V and V2I settings to
illustrate representation quality and retrieval performance.

\textbf{Feature Embedding Analysis.} 
Fig.~\ref{tsne} (a–d) shows t-SNE visualizations under different training stages.
In Fig.~\ref{tsne} (a), features from the pretrained encoder exhibit severe identity entanglement and a pronounced visible–infrared separation, indicating strong modality bias.
Fig.~\ref{tsne} (b) shows that baseline intra-modality clustering improves identity separability within each modality, yet a large modality gap persists.
With the introduction of CIW in Fig.~\ref{tsne} (c), visible and infrared features become significantly closer, demonstrating improved modality invariance at an early stage.
Finally, Fig.~\ref{tsne} (d) shows that PGUR further refines cross-modality alignment, yielding well-separated identity clusters with better interleaving across modalities.

\textbf{Distance Distribution Analysis.}  
Fig.~\ref{dis} (a–d) illustrates intra-class and inter-class distance distributions under different settings.
In Fig.~\ref{dis} (a), the two distributions heavily overlap, indicating weak identity discriminability.
The baseline in Fig.~\ref{dis} (b) partially enlarges the separation, yet the margin remains limited due to modality discrepancies.
With CIW in Fig.~\ref{dis} (c), the distance gap is substantially increased, reflecting improved identity consistency.
The full CBA framework in Fig.~\ref{dis} (d) achieves the largest margin, demonstrating the most discriminative and well-aligned representations.

\textbf{Retrieval Results Analysis.}  
Fig.~\ref{retrieval} presents qualitative cross-modality retrieval results under
I2V and V2I settings for three representative queries.
The baseline (B/L) exhibits severe cross-modality failure, where most top-ranked
results (Rank@1–6) correspond to incorrect identities, indicating that modality
bias dominates the learned representation.
After introducing CIW (B/L+CIW), correct matches (green boxes) consistently
appear at higher ranks, demonstrating that causal intervention effectively
enhances early identity awareness and suppresses modality-specific shortcuts.
With the full CBA framework, correct identities dominate the top-ranked results
across all queries, while false positives are largely eliminated, reflecting
more reliable and stable cross-modality alignment enabled by PGUR.

\section{Conclusion}
In this paper, we proposed Causal Bootstrapped Alignment (CBA), a novel framework for USL-VVI-ReID.
By analyzing generic pre-trained encoders from a causal and structural perspective, we identify two fundamental bottlenecks: severe intra-modality identity confusion and cross-modality clustering granularity mismatch.
To address the former, we introduce Causal Intervention Warm-up (CIW), which
performs sequence-level causal interventions by exploiting temporal identity
consistency and perturbation-induced cross-modality identity invariance, suppressing modality- and motion-related confounders while preserving identity-discriminative semantics. To resolve the latter, we propose Prototype-Guided Uncertainty Refinement (PGUR), which alleviates one-to-many cross-modal matching ambiguity via coarse-to-fine structural refinement, leveraging reliable visible prototypes to reorganize infrared clusters with uncertainty-aware supervision. Extensive experiments on the HITSZ-VCM and BUPTCampus benchmarks demonstrate the effectiveness and robustness of the proposed framework.

\bibliography{extracted}

\end{document}